\documentclass{article}

\usepackage{microtype}
\usepackage{graphicx}
\usepackage{subcaption}
\usepackage{booktabs} 

\usepackage[preprint]{icml2026}

\usepackage{amsmath}
\usepackage{amssymb}
\usepackage{mathtools}
\usepackage{amsthm}

\usepackage[textsize=tiny]{todonotes}

\usepackage{xcolor}
\definecolor{yvesblue}{RGB}{0, 47, 167}
\usepackage[hidelinks,
colorlinks,
urlcolor = red!88!black,
citecolor = yvesblue,]{hyperref}
\usepackage{setspace} 
\usepackage{multirow}
\usepackage{algorithm}
\usepackage{algpseudocode}
\usepackage{algcompatible}
\usepackage{tabularx}
\usepackage{makecell}
\usepackage{ltablex}
\usepackage{threeparttable}
\usepackage{caption}
\usepackage{subcaption}
\usepackage{graphicx}
\usepackage{pgf,tikz}
\usepackage{wrapfig}
\usetikzlibrary{positioning, calc, arrows.meta}

\usepackage{amsmath}
\usepackage{amssymb}
\usepackage{mathtools}
\usepackage{amsthm}
\makeatletter
\newtheoremstyle{define}%
  {3pt}{3pt}{}{}
  {\bfseries}{}
  {0pt}{\thmname{#1} \thmnumber{#2} \normalfont\thmnote{ (#3)}. }
\makeatother
\theoremstyle{define}

\usepackage{etoc}
\etocdepthtag.toc{mtchapter}
\etocsettagdepth{mtchapter}{subsection}
\etocsettagdepth{mtappendix}{none}

\usepackage[table]{xcolor}
\definecolor{gray}{rgb}{0.5,0.5,0.5}
\definecolor{light-gray}{gray}{0.2}
\definecolor{dark-gray}{gray}{0.15}
\definecolor{brickblack}{rgb}{0.0, 0.0, 0.0}
\definecolor{brickred}{rgb}{0.8, 0.25, 0.33}
\definecolor{brickblue}{rgb}{0.432, 0.600, 0.793}

\usepackage{amssymb, amsmath, amsthm}
\usepackage{bm}
\usepackage{thmtools, thm-restate}
\declaretheoremstyle[
  spaceabove=3pt,
  spacebelow=3pt,
  bodyfont=\normalfont,
  headfont=\bfseries,
  notefont=\normalfont,
  headpunct={ },
  postheadspace=0pt,
  headformat=\bfseries\NAME\ \NUMBER\ \NOTE\normalfont.
]{define2}
\declaretheoremstyle[
  spaceabove=3pt,
  spacebelow=3pt,
  bodyfont=\normalfont,
  headfont=\bfseries,
  notefont=\normalfont,
  headpunct={},
  postheadspace=0pt,
  headformat=\bfseries\NAME\ \NUMBER\ \NOTE\normalfont
]{define3}

\usepackage{enumitem}

\usepackage{arydshln}
\usepackage{booktabs}

\usepackage{pgfplots}
\usepackage{pdfpages}

\usepackage[export]{adjustbox} 
\usepackage{array}

\usepackage{changepage}

\usepackage{arydshln}

\usepackage{placeins}
\usepackage{afterpage}

\usepackage{tcolorbox}
\definecolor{darkerlogocolor}{RGB}{20, 0, 145}  
\newtcolorbox{ttcolorbox}[1][]{
  colframe=darkerlogocolor,
  colback=darkerlogocolor!4!white,
  title=#1,
  left=0.5mm,
  right=1mm,
  top=-2mm,
  bottom=-3mm
}

\definecolor{darkergreencolor}{RGB}{
40, 120, 45}
\newtcolorbox{greenttbox}[1][]{
  colframe=darkergreencolor,
  colback=darkerlogocolor!4!white,
  title=#1,
  left=0.5mm,
  right=1mm,
  top=-2mm,
  bottom=-3mm
}

\usepackage{listings}
\lstdefinestyle{mystyle}{
    commentstyle=\color{codegreen},
    keywordstyle=\color{magenta},
    numberstyle=\tiny\color{codegray},
    stringstyle=\color{codepurple},
    basicstyle=\ttfamily\tiny,
    breakatwhitespace=false,         
    breaklines=true,                 
    captionpos=b,                    
    keepspaces=true,                                   
    numbersep=5pt,                  
    showspaces=false,                
    showstringspaces=false,
    showtabs=false,                  
    tabsize=2
}
\makeatletter
\AtBeginDocument{
\renewcommand{\sectionautorefname}{\S\@gobble}

\renewcommand{\subsectionautorefname}{\S\@gobble} 
\renewcommand{\subsubsectionautorefname}{\S\@gobble}
\renewcommand{\appendixautorefname}{\S\@gobble}
}
\makeatother

\icmltitlerunning{TabRAG: Improving Tabular Document Question Answering for RAG via Structured Representations}

\begin{document}

\twocolumn[
  \icmltitle{TabRAG: Improving Tabular Document Question Answering \\for Retrieval Augmented Generation via Structured Representations}



  \icmlsetsymbol{equal}{*}




  \begin{icmlauthorlist}
    \icmlauthor{Jacob Si}{equal,imperial}
    \icmlauthor{Mike Qu}{equal,columbia}
    \icmlauthor{Michelle Lee}{imperial}
    \icmlauthor{Marek Rei}{imperial}
    \icmlauthor{Yingzhen Li}{imperial}
  \end{icmlauthorlist}

  \icmlaffiliation{imperial}{Imperial College London}
  \icmlaffiliation{columbia}{Columbia University}

  \icmlcorrespondingauthor{Jacob Si}{y.si23@imperial.ac.uk}

  \icmlkeywords{tabular RAG, retrieval-augmented generation, tabular data}

  \vskip 0.3in
]



\printAffiliationsAndNotice{\icmlEqualContribution}

\begin{abstract}
    Incorporating external knowledge bases in traditional retrieval-augmented generation (RAG) relies on parsing the document, followed by querying a language model with the parsed information via in-context learning. While effective for text-based documents, question answering on tabular documents often fails to generate plausible responses. Standard parsing techniques lose the two-dimensional structural semantics critical for cell interpretation. In this work, we present \textsc{TabRAG}, a parsing-based RAG framework designed to improve tabular document question answering via structured representations. Our framework consists of layout segmentation that decomposes the document inputs into a series of components, enabling fine-grained extraction. Subsequently, a vision language model parses and extracts the document tables into a hierarchically structured representation. In order to cater various table styles and formats, we integrate a self-generated in-context learning module that guides the table extraction process. Experimental results demonstrate that \textsc{TabRAG} outperforms existing popular parsing techniques across a broad suite of evaluation and ablation benchmarks.
    Code is available at: \url{https://github.com/jacobyhsi/TabRAG}.
\end{abstract}

\begin{figure*}[!t]
    \centering
    \includegraphics[width=\textwidth]{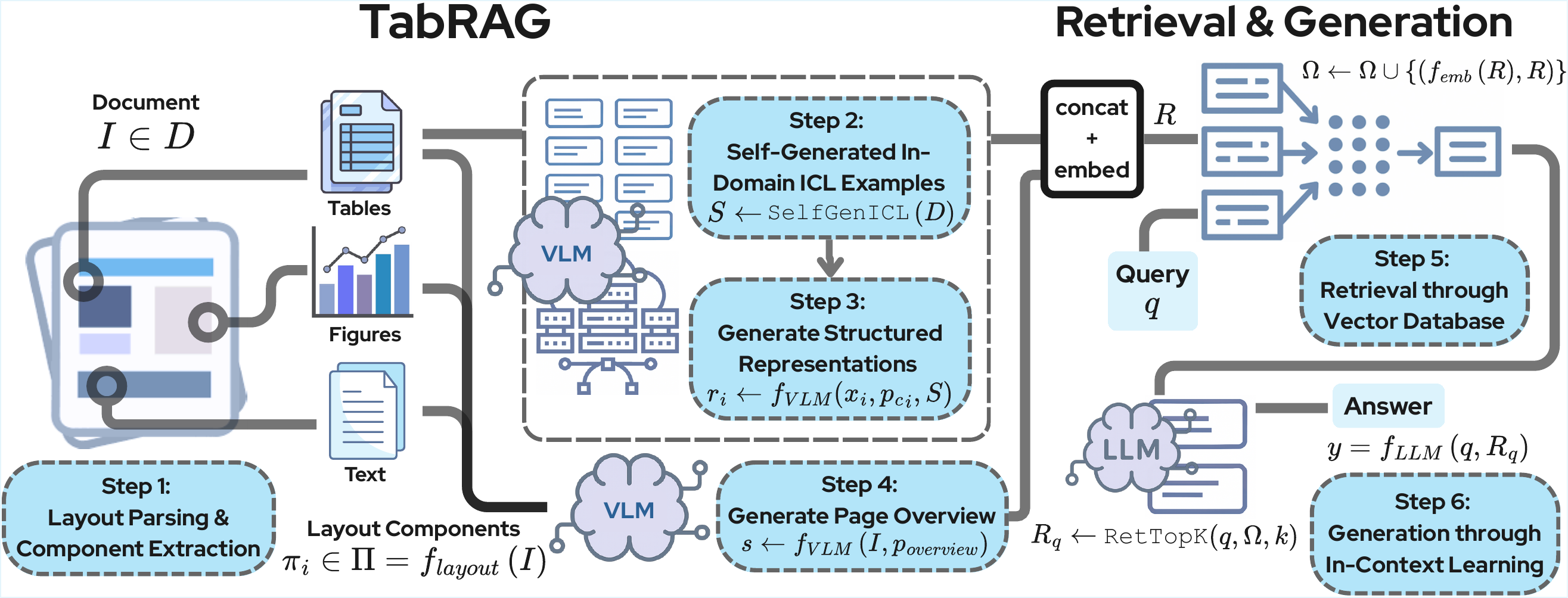}
    \caption{The \textsc{TabRAG} Architecture, a parsing-based RAG pipeline designed specifically for tables. A layout detection model is applied to segment various components from the documents. Specifically, the tables are then passed into a vision language model, which extracts cell values along with their corresponding column and row names in a structured representation. The structured representation is then coupled with self-generating in-context learning examples that guide its output.}
    \label{fig:tabrag_architecture}
    \vspace{-3mm}
\end{figure*}

\section{Introduction}
Tabular documents are the standard for organizing actionable data in professional and research environments \cite{vanbreugel2024tabularfoundationmodelsresearch, hollmann2023tabpfntransformersolvessmall, si2024interpretabnetdistillingpredictivesignals, arik2020tabnetattentiveinterpretabletabular, si2025tabrep}. These documents are used to record and communicate critical information across various domains, from auditing financial statements in corporate settings to personal day-to-day tax filings \cite{tatdqaZhu_2022, kim2024tablevqabench, zhao2024tabpedia_comtqa}. Consequently, effective table retrieval and understanding are essential for accurate data analysis and decision-making. Unlike plain text, which follows a natural sequence, a notable feature unique to tabular data is that it conveys information through a two-dimensional layout, where the value of each cell depends on both its row and column context, along with added nuances such as units and table structure.

Retrieval-Augmented Generation (RAG) has become an established framework that enhances large language models by grounding their generation in external knowledge sources \cite{lewis2021retrievalaugmentedgenerationknowledgeintensivenlp, guu2020realmretrievalaugmentedlanguagemodel, wei2025instructraginstructingretrievalaugmentedgeneration}. In real-world industries, RAG is widely used in settings such as personalized enterprise search, and question answering over private documents,  many of which involve large, complex tables \cite{OpenAIBusiness, Cohere, llamaindex}. Traditional RAG \cite{lewis2021retrievalaugmentedgenerationknowledgeintensivenlp, guu2020realmretrievalaugmentedlanguagemodel} relies on parsing documents to obtain text, followed by querying a large language model (LLM) with the corresponding retrieved information via in-context learning \cite{llamaindex, langchain}. While straightforward, traditional RAG pipelines often struggle to generate plausible answers when queried about tabular documents \cite{shaikh2025whyragbreaks, kylo_fromgistr_2025_ragtables}. During inference, the retrieved tables are typically serialized into linear text before provided as in-context learning examples. However, this conversion process creates a bottleneck as the resulting text-based representations fail to effectively mirror the original two-dimensional dependencies. The complicated table styles and formatting overwhelm the LLM’s reasoning abilities and lead to hallucinated or imprecise answers. The best performing parsers that serialize the data either lack customizability and can only perform flat-text extraction \cite{pymupdf, pytesseract} or require careful prompt engineering \cite{qwen3technicalreport, openai2025gpt52, gemma2025gemma3}, which we tackle in our paper.

Since tabular documents are inherently structured and visually grounded, effectively integrating them into RAG pipelines requires more than typical extraction methods. In this work, we propose \textsc{TabRAG}, a parsing-based RAG framework that improves tabular document question answering via structured representations. \textsc{TabRAG} first segments the document via a layout detector \cite{li2022ditselfsupervisedpretrainingdocument} into various components. This produces smaller chunks of data that allow the pipeline to focus on localized content while ensuring each chunk fits within a model’s context window. Next, we engineer prompts for a vision language model (VLM) to parse various component types \cite{brown2020languagemodelsfewshotlearners, sahoo2025systematicsurveypromptengineering}. Specifically, for table components, we focus on extracting the values and units, along with their corresponding column and row names, into a structured representation. We then include an autonomous in-context learning module that self-generates exemplars, serving as supervision by facilitating domain knowledge for the VLM during table extraction. We conduct comprehensive experiments against parsing baselines across various datasets and benchmarks. The results demonstrate that \textsc{TabRAG} outperforms existing methods in generation while remaining competitive for retrieval. To the
best of our knowledge, the proposed work is the first to explore the design of parsed representations for tabular question answering tasks in RAG. Our architecture can be found in Figure \ref{fig:tabrag_architecture}.

\section{Method}\label{sec:method}

Given a dataset $D$ that contains $I$ documents, each of which constitutes a single page, we consider various parsing strategies that extract its contents. In the following section, we introduce how we construct our \textsc{TabRAG} framework, starting by decomposing our documents into focused tasks (\autoref{method:layout_segmentation}), followed by the design of our structured representation (\autoref{method:structured_rep}). Lastly, we augment our structured representation with an autonomous self-generated in-context learning (ICL) example (\autoref{method:self-gen}), and concluded with the retrieval and generation pipeline (\autoref{method:retrieval_and_generation}).


\subsection{Layout Segmentation} 
\label{method:layout_segmentation}

Layout segmentation identifies and segments visually coherent regions to preserve the document's structural organization. Given an input document page $I$, we introduce a layout detection model $f_{\text{layout}}$ based on the Document Image Transformer  \cite{li2022ditselfsupervisedpretrainingdocument}, a self-supervised transformer model for document image understanding trained on large-scale datasets such as PubLayNet \cite{zhong2019publaynetlargestdatasetdocument}. The layout model serves to predict bounding box positions and semantic labels for document components. More formally, 
\begin{equation}
   \Pi \leftarrow f_{\text{layout}}(I), \quad \text{where } \Pi = \{\pi_i\}_{i=1}^{M}, \ \pi_i = (b_i, x_i, c_i),
\end{equation}
and $(b_i, x_i, c_i)$ represents a tuple of bounding box coordinates, cropped image, and component type. The component type $c_i \in \{0, 1, 2, 3\}$ maps to a semantic label in the set $\{\text{\texttt{title}}, \text{\texttt{text}}, \text{\texttt{figure}}, \text{\texttt{table}}\}$. Layout segmentation allows for explicit localized reasoning and addresses several key limitations of page-level methods. When an entire page is represented as a single visual embedding, global attention blurs fine-grained dependencies \cite{liu2023lostmiddlelanguagemodels} that are especially crucial in tables, such as column alignments, header associations, and cell-to-cell relations. A nice byproduct of layout segmentation is that it also reduces the effective context length, allowing the input to fit within the context window of the VLMs.

In contrast to standard document understanding models, such as LayoutLMv3 \cite{huang2022layoutlmv3pretrainingdocumentai} and DocFormer \cite{appalaraju2021docformerendtoendtransformerdocument}, that use spatial embeddings to correlate text with 2D coordinates and layout semantics, \textsc{TabRAG} models these dependencies explicitly through region-level segmentation, ensuring that relationships are preserved in the representation itself. This allows \textsc{TabRAG} to establish a more faithful and structurally grounded foundation for downstream multimodal processing. 

\begin{algorithm*}[!t]
    \small
    \setstretch{1}
    \caption{Self-Generated In-Context Learning}
    \label{alg:icl_sampling}
    \begin{algorithmic}[1]
        \STATE \textbf{Input:} Dataset $\mathcal{D}$, VLM $f_{\text{VLM}}$, tokenizer $f_{\text{tok}}$, target examples $K{=}3$, candidate pages $M{=}20$, min tokens $L_{\min}$, max tokens $L_{\max}$
        
        \STATE \textbf{Output:} Selected ICL examples $\mathcal{S}$

        \STATE $\mathcal{I} \leftarrow \textsc{CollectPage}(\mathcal{D})$ \hfill \textcolor{gray}{\Comment{accumulate document pages in the dataset}}
        
        \STATE $\tilde{\mathcal{I}} \leftarrow \textsc{Shuffle}(\mathcal{I})$ \hfill \textcolor{gray}{\Comment{shuffle to ensure fairness}}
        
        \STATE $\mathcal{T}, \mathcal{S} \leftarrow \emptyset, \emptyset$ \hfill \textcolor{gray}{\Comment{store largest tables of all $\tilde{\mathcal{I}}$ and ICL examples}}

        \FORALL{$I \in \tilde{\mathcal{I}}$ \textbf{and} $|\mathcal{T}| < M$}
        
            \STATE $\Pi \leftarrow f_{\text{layout}}(I)$ \hfill \textcolor{gray}{\Comment{where $\Pi = \{\pi_i\}_{i=1}^{M}$}} 
            
            \STATE $\Pi^{\text{tbl}} \leftarrow \{\pi_i \in \Pi \mid c_i==\texttt{table}\}$ \hfill \textcolor{gray}{\Comment{$\pi_i = (b_i, x_i, c_i)$ = (bbox, cropped img, component type)}}
            
            \IF{$\Pi^{\text{tbl}} \neq \emptyset$}

                
                \STATE $\mathcal{T} \leftarrow \mathcal{T} \cup \{\arg\max_{\pi \in \Pi^{\text{tbl}}} \textsc{Area}(b_i)\}$ \hfill \textcolor{gray}{\Comment{extract largest table from the extracted tables}}
            \ENDIF
        \ENDFOR

        \STATE $\mathcal{T} \leftarrow \textsc{SortByLargestArea}(\mathcal{T})$

        \FORALL{$x \in \mathcal{T}$ \textbf{and} $|\mathcal{S}| < K$} \hfill \textcolor{gray}{\Comment{generate ICL examples from the top $K$ largest tables}}
        
            \STATE $m \leftarrow f_{\text{VLM}}(x,p_{\text{markdown}})$ \hfill \textcolor{gray}{\Comment{markdown structured representation}}
            
            \STATE $j \leftarrow f_{\text{VLM}}(x,p_{\text{json}})$ \hfill \textcolor{gray}{\Comment{json structured representation}}
            
            \STATE $L \leftarrow f_{\text{tok}}(m) + f_{\text{tok}}(j)$
            
            \IF{$L_{\min} \le L < L_{\max}$} \hfill \textcolor{gray}{\Comment{ensure validity of representation}}
                \STATE $\mathcal{S} \leftarrow \mathcal{S} \cup \{(m, j)\}$
            \ENDIF
        \ENDFOR

        \STATE \textbf{return} $\mathcal{S}$
    \end{algorithmic}
\end{algorithm*}

\subsection{Structured Representation}
\label{method:structured_rep}
Once the document is decomposed into localized regions, each segment is semantically interpreted using a vision language model. For each image $x_i$, we prompt a VLM $f_\text{VLM}$ with a component-specific prompt $p_{c_i}$ to translate the visual structure into a concise, structured, yet information-dense representation of each region: 
\begin{equation}
    r_i \leftarrow f_{\text{VLM}}(x_i, p_{c_i}).
\end{equation}
The VLM is instructed to extract texts and titles directly, while providing detailed interpretations and descriptions for figures. Notably, for tables, we engineer a prompt for the VLM to generate a JSON-based structured representation. Formally, we first provide our VLM with the role of an ``information extraction engine'' \cite{kong2024betterzeroshotreasoningroleplay} tasked with extracting visible cells in the input table image into JSON objects, where each table cell must be represented with its corresponding row name, column name, value, and unit. We then define rules addressing nuances, including punctuation, normalizations of symbols and numbers, table formatting such as multi-level text and headers, as well as empty cells. To further guide the extraction process, we couple our instructions with self-generated ICL examples, creating a system agnostic to diverse table styles. 

Instead of performing flat-text extraction on the table, the VLM is used to provide a structured understanding of the visual elements. We tune the VLM's prompt to capture cues such as alignment, cell grouping, and formatting style, corresponding to how information is organized and interpreted by humans. This semantic enrichment transforms raw visual segments into a coherent textual representation that can later be reasoned over by a language model.


In the case where the layout model fails, either due to the model's lack of generalization or ambiguous visual boundaries, we introduce a fallback mechanism in which the vision-language model is also applied to the full document image: $f_{\text{VLM}}(I,p_{\texttt{page}})$. This full-page inference provides a coarse global description that captures the overall content, even when region extraction is unreliable. During retrieval, summaries produced under this fallback mode are treated equivalently to region-level descriptions, allowing the system to maintain continuity and prevent failure propagation across the pipeline. 


\subsection{Self-Generated In-Context Learning}
\label{method:self-gen}

In-context learning has been established as an emergent property of large autoregressive language models \cite{brown2020languagemodelsfewshotlearners}. Language models demonstrate reasoning abilities from few-shot or chain-of-thought demonstrations \cite{wei2023chainofthoughtpromptingelicitsreasoning, wang2023selfconsistencyimproveschainthought}, improving overall coherence in their outputs. Considering that tables are formatted in varying complex styles (see Appendix \ref{app:datasets}), supervision is necessary to guide the VLM in its extraction. Given a VLM, a natural way to incorporate ICL could be to input example image tables along with instructions. However, tokenizing images is costly and consumes a significant amount of the context window. Therefore, the architecture necessitates a text-based ICL example. Inspired by recent agentic workflows \cite{yao2022react, shinn2023reflexionlanguageagentsverbal}, we design a VLM-based autonomous module that self-generates text-based ICL examples, supplementing our prompt for extracting structured table representations.

\begin{figure*}[!t]
    \centering
    \includegraphics[width=0.9\textwidth]{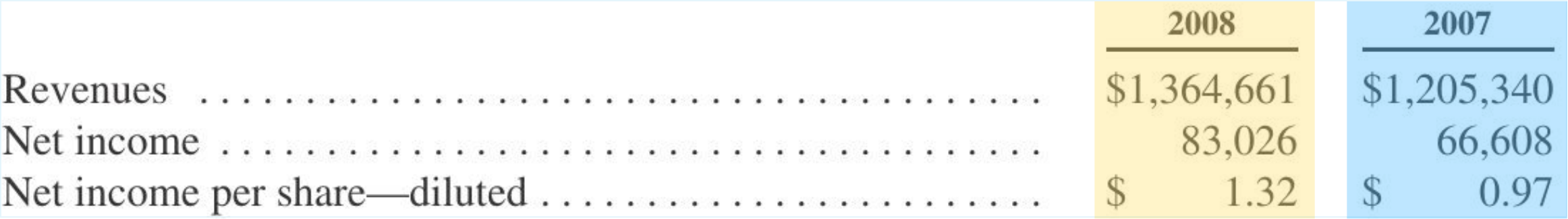}
    \begin{ttcolorbox}[\textsc{TabRAG}'s Structured Representation]
    \begin{minipage}{\linewidth}
    \begin{lstlisting}[basicstyle=\scriptsize\ttfamily,escapeinside={(*@}{@*)}]
(*@\colorbox{yellow!30}{\{"row": "Revenues", "column": "2008", "value": "\$1,364,661", "units": "\$"\}}@*)
(*@\colorbox{cyan!20}{\{"row": "Revenues", "column": "2007", "value": "\$1,205,340", "units": "\$"\}}@*)
(*@\colorbox{yellow!30}{\{"row": "Net income", "column": "2008", "value": "83,026", "units": null\}}@*)
(*@\colorbox{cyan!20}{\{"row": "Net income", "column": "2007", "value": "66,608", "units": null\}}@*)
(*@\colorbox{yellow!30}{\{"row": "Net income per share-diluted", "column": "2008", "value": "\$ 1.32", "units": "\$"\}}@*)
(*@\colorbox{cyan!20}{\{"row": "Net income per share-diluted", "column": "2007", "value": "\$ 0.97", "units": "\$"\}}@*)
    \end{lstlisting}
    \end{minipage}
    \end{ttcolorbox}

    \vspace{-3mm}

    \begin{greenttbox}[LLM Query Engine (Input/Output)]
    \begin{minipage}{\linewidth}
    \begin{lstlisting}[basicstyle=\scriptsize\ttfamily,escapeinside={(*@}{@*)}]
You are a helpful assistant. Use the information from the documents below to answer the question.

{`TabRAG's Structured Representation'}

Question: {`What is the Net income per share--diluted in 2007?'} 

Answer: $ 0.97 
    \end{lstlisting}
    \end{minipage}
    \end{greenttbox}
    
    \vspace{-2mm}
    \caption{Qualitative example from TableVQA of \textsc{TabRAG} aiding the LLM Query Engine in delivering the correct output.}
    \label{fig:qualitative_example}
    \vspace{-2mm}
\end{figure*}

We formalize this process as a multi-stage selection and generation pipeline. Given a dataset $\mathcal{D}$, we first accumulate and shuffle a set of document pages $\tilde{\mathcal{I}}$ to ensure fairness. For each page, we apply a layout segmentation model $f_{\text{layout}}$ and extract layout components $\Pi$. We then filter specifically for elements identified as a table:
\begin{equation}
    \Pi^{\text{tbl}} \leftarrow \{\pi_i \in \Pi \mid c_i=\texttt{table}\}.
\end{equation}
To ensure that the generated ICL examples are sufficiently complex, we select the largest table from each page based on its bounding box area: $\arg\max_{\pi \in \Pi^{\text{tbl}}} \textsc{Area}(b_i)$. The $M$ candidate tables with the largest area are collected, then each processed by a VLM to generate dual structured representations: a markdown string $m$ and a JSON object $j$:
\begin{align}
    m &\leftarrow f_{\text{VLM}}(x,p_{\text{markdown}}), \\
    j &\leftarrow f_{\text{VLM}}(x,p_{\text{json}}).
\end{align}
To ensure the validity of the demonstration, we compute the total token length $L$:
\begin{equation}
    L \leftarrow f_{\text{tok}}(m) + f_{\text{tok}}(j),
\end{equation}
and enforce a constraint such that $L$ is at least $L_{\min}$ and at most $L_{\max}$ tokens in length: $L_{\min} \le L < L_{\max}$. The markdown and JSON are then concatenated to form the final input-output pairs and accumulated in $\mathcal{S}$ until it reaches the desired $K$ target ICL examples. $\mathcal{S}$ can now be used as input to the VLM when generating the structured representations:
\begin{equation}
    r_i \leftarrow 
    \begin{cases} 
        f_{\text{VLM}}(x_i, p_{c_i}, \mathcal{S}) & \text{if } c_i = \texttt{table} \\
        f_{\text{VLM}}(x_i, p_{c_i}) & \text{otherwise}
    \end{cases}
    \label{eq:structured_rep_final}
\end{equation}
Our self-generated ICL algorithm and its recommended hyperparameters can be found in Algorithm \ref{alg:icl_sampling}. This filtering mechanism ensures that the selected set $\mathcal{S}$ consists of $K$ high-fidelity examples that fit within the model's context window. Most importantly, the algorithm incorporates the topological complexities of the table's design and format, as well as its domain-specific semantics. To wrap up our method, a qualitative example is displayed in Figure \ref{fig:qualitative_example}.

\subsection{Retrieval and Generation}
\label{method:retrieval_and_generation}

After yielding the representations across the whole dataset, $\mathcal{R}$, they are then embedded into a high-dimensional latent space, $f_{emb}(\mathcal{R})$ \cite{zhang2025qwen3embeddingadvancingtext}, and indexed within a vector store, $\Omega$ \cite{douze2025faisslibrary}. At inference time, a similarity search is then performed against the vector store to retrieve the $k$ most relevant documents to the query, $q$. The retrieved document is then used as an in-context learning example for an LLM to generate an answer to the query.


We provide a summary of our Methods section (\autoref{sec:method}) in the appendix. Our end-to-end \textsc{TabRAG} algorithm can be found in Appendix \ref{app:algorithm}. Prompts for \textsc{TabRAG}'s self-generated ICL examples and \textsc{TabRAG}'s structured representations can be found in Appendix \ref{app:prompts_selfgen-icl} and \ref{app:prompts_tabrag}. \textsc{TabRAG}'s stage-by-stage outputs can be found in Appendix \ref{app:output_layout_detection} and \ref{app:output_tabrag}.





\begin{table*}[!t]
    \centering
    \caption{Overall generation performance in accuracy (\%). We assume that the query corresponds to the ground truth document. In an in-context learning setting, we query an LLM to use the representations synthesized by different parsing methods to answer the question. To evaluate, we use exact-match accuracy, which determines whether the ground-truth answers are included in the model generations.}
    \label{tbl:generation}
    \vspace{-2mm}
    \begin{normalsize}
    \begin{threeparttable}
    \begin{sc}
    \resizebox{\textwidth}{!}{
    \begin{tabular}{lcccccc}
    \toprule
    & \multicolumn{6}{c}{Accuracy (\%) $\uparrow$} \\
    \cmidrule(lr){2-7}
    Models & TAT-DQA & MP-DocVQA & WikiTQ & TableVQA  & ComTQA & Average \\
    \midrule
    PyMuPDF & 66.83{\tiny$\pm$.000} & N/A & 59.49{\tiny$\pm$.004} & N/A & N/A & 63.16 \\
    PyTesseract & 62.01{\tiny$\pm$.002} & 59.73{\tiny$\pm$.002} & 49.32 {\tiny$\pm$.005} & 51.88{\tiny$\pm$.001} & 40.90{\tiny$\pm$.001} & 52.77 \\
    Deepseek-OCR & 64.45{\tiny$\pm$.002}  & 73.47{\tiny$\pm$.003} &  60.40{\tiny$\pm$.002} & 68.57{\tiny$\pm$.002} & 53.30{\tiny$\pm$.005} & 64.04 \\
    
    Gemma-3-27b-it & 43.92{\tiny$\pm$.002} & 69.08{\tiny$\pm$.003} &  56.75{\tiny$\pm$.002} & 64.75{\tiny$\pm$.003} & 51.70{\tiny$\pm$.002} & 57.24 \\
    
    Qwen3-VL-8B-Instruct & 66.26{\tiny$\pm$.001} & 85.81{\tiny$\pm$.002} & 61.71{\tiny$\pm$.005} & 66.29{\tiny$\pm$.002} & 67.80{\tiny$\pm$.001} & 69.57 \\

    Qwen3-VL-32B-Instruct & 66.56{\tiny$\pm$.003} & 86.32{\tiny$\pm$.003} & 65.95{\tiny$\pm$.007} & 66.07{\tiny$\pm$.001} & 71.30{\tiny$\pm$.002} & 71.24 \\

    GPT-5.2 & 68.54{\tiny$\pm$.001} & 86.26{\tiny$\pm$.001} & 62.23{\tiny$\pm$.002} & 67.40{\tiny$\pm$.002} & 73.90{\tiny$\pm$.003} & 71.67 \\

    \midrule
    \rowcolor{gray!20}\cellcolor{white}TabRAG$_\text{8B}$ & 92.56{\tiny$\pm$.002} &  88.87{\tiny$\pm$.003} &  68.49{\tiny$\pm$.003} &  70.60{\tiny$\pm$.004} & \textbf{82.60{\tiny$\pm$.002}} & \textbf{80.62} \\
    \rowcolor{gray!20}\cellcolor{white}TabRAG$_\text{32B}$ & \textbf{92.70{\tiny$\pm$.002}} & \textbf{88.93{\tiny$\pm$.002}} & \textbf{71.68{\tiny$\pm$.004}} & \textbf{71.20{\tiny$\pm$.001}} & 75.80{\tiny$\pm$.002} & 80.06 \\

    \bottomrule
    
    \end{tabular}}
    \end{sc}
    \end{threeparttable}
    \end{normalsize}
    \vspace{-2mm}
\end{table*}
\section{Experiments}\label{sec:4_experiments}

\subsection{Setup}

\textsc{TabRAG} employs Qwen3-VL-8B-Instruct (\textsc{TabRAG$_{\text{8B}}$}) and Qwen3-VL-32B-Instruct (\textsc{TabRAG$_{\text{32B}}$}) as the vision-language model \cite{qwen3technicalreport}. We use Qwen3-Embedding-8B \cite{zhang2025qwen3embeddingadvancingtext} as the embedding backbone for retrieval, where the embeddings are stored in a FAISS vector store \cite{douze2025faisslibrary}. Lastly, a Qwen3-8B (No Think) language model is used as the downstream query engine. In our main text, we focus on \textsc{TabRAG$_{\text{8B}}$} for our ablation studies. Experimental results are obtained over an average of 3 evaluation seeds. Document pages used in our self-generated ICL are excluded from evaluation. Further details on the setup can be found in Appendix \ref{app:implementation}.

\textbf{Datasets}. We evaluate \textsc{TabRAG} across document question-answering benchmarks that include tables and vary in layout complexity and domain coverage. These include TAT-DQA \cite{tatdqaZhu_2022} and TableVQA-Bench (TableVQA) \cite{kim2024tablevqabench}, which focus on real-world financial reports; ComTQA \cite{zhao2024tabpedia_comtqa}, which examines a combination of financial reports and scientific articles; MP-DocVQA \cite{mpdocvqatito2023hierarchicalmultimodaltransformersmultipage}, which targets multipage industry scanned documents; and WikiTableQuestions (WikiTQ) \cite{wikitqpasupat2015compositionalsemanticparsingsemistructured}, which contains semi-structured HTML tables. To ensure consistency across modalities, all multi-page documents are preprocessed into single-page units and converted into high-resolution PNG images for methods requiring visual input. See Appendix \ref{app:datasets} for more details. 

\textbf{Baselines}. We compare \textsc{TabRAG} against existing parsing frameworks: PyMuPDF \cite{pymupdf}, PyTesseract \cite{pytesseract}, Deepseek-OCR \cite{wei2025deepseekocr}, and VLMs including Gemma-3-27B-it \cite{gemma2025gemma3}, Qwen3-VL-8B/32B-Instruct \cite{qwen3technicalreport} and GPT-5.2 \cite{openai2025gpt52}, which represent text-based parsing, OCR-based extraction, and transformer-based VLMs. See Appendix \ref{app:baselines} for more details. The prompts used for our baseline VLM experiments are shown in Appendix \ref{app:prompts_baselines}. While these parsers are highly accurate in extraction quality, they are limited by the linearized nature of their output. Specifically, PyMuPDF and PyTesseract follow a raster scan order, while VLMs typically default to markdown serialization. Both approaches discard the table's spatial hierarchy by collapsing two-dimensional dependencies into a one-dimensional sequence.

\textbf{Benchmarks}. To evaluate generation performance, we use exact-match accuracy (excluding whitespace, punctuation, and capitalization), which determines whether the ground-truth answers are included in the model generations \cite{wei2025instructraginstructingretrievalaugmentedgeneration, mallen2023not, schick2023toolformer}. For retrieval, we use mean reciprocal rank (MRR@10) \cite{yu2025visragvisionbasedretrievalaugmentedgeneration, masry2022chartqa, methani2020plotqa}, computed as the inverse of the rank of the first relevant document among the top-10 retrieved candidates, and averaged across all queries, assigning higher scores to relevant documents retrieved at earlier positions. Further details can be found Appendix \ref{app:benchmarks}. The prompt used for our query engine can be found in Appendix \ref{app:prompts_llm_query_engine}.

\begin{table*}[!t]
    \centering
    \caption{Overall retrieval performance in mean reciprocal rank (MRR@10). Given a query, the model searches over ten indexed document representations and ranks them based on semantic similarity. Note that since we emphasize on generation, an optimal method of deploying \textsc{TabRAG} is to combine it with state-of-the-art retrievers and leverage the corresponding \textsc{TabRAG} vector stores during generation.}
    \label{tbl:reject_answering}
    \vspace{-2mm}
    \begin{normalsize}
    \begin{threeparttable}
    \begin{sc}
    \resizebox{\textwidth}{!}{
    \begin{tabular}{lcccccc}
    \toprule
    & \multicolumn{6}{c}{MRR@10 $\uparrow$} \\
    \cmidrule(lr){2-7}
    Models & TAT-DQA & MP-DocVQA & WikiTQ & TableVQA & ComTQA & Average \\
    \midrule
    PyMuPDF & 75.60{\tiny$\pm$.001} & N/A & \textbf{86.74{\tiny$\pm$.001}} & N/A & N/A & 81.17 \\
    PyTesseract & \textbf{75.95{\tiny$\pm$.001}} & 83.77{\tiny$\pm$.001} & 82.23{\tiny$\pm$.003} & 85.48{\tiny$\pm$.004} & 81.36{\tiny$\pm$.002} & 81.76 \\
    Deepseek-OCR & 73.66{\tiny$\pm$.002}  & 84.05{\tiny$\pm$.002}  &  82.01{\tiny$\pm$.003} & 88.72{\tiny$\pm$.002} & 80.25{\tiny$\pm$.005} & 81.74 \\
    gemma-3-27b-it & 71.31{\tiny$\pm$.004}  & 84.48{\tiny$\pm$.001}&  85.81{\tiny$\pm$.002} & 88.75{\tiny$\pm$.005} & 80.77{\tiny$\pm$.002} & 82.22 \\
    Qwen3-VL-8B-Instruct & 73.88{\tiny$\pm$.002} & 84.59{\tiny$\pm$.002} & 86.20{\tiny$\pm$.001} & 88.57{\tiny$\pm$.002} & 82.34{\tiny$\pm$.001} & 83.12 \\
    Qwen3-VL-32B-Instruct & 73.95{\tiny$\pm$.002} & 84.69{\tiny$\pm$.003} & 86.24{\tiny$\pm$.001} & 88.16{\tiny$\pm$.001} & \textbf{82.59{\tiny$\pm$.001}} & 83.13 \\
    GPT-5.2 & 74.23{\tiny$\pm$.002} & 84.69{\tiny$\pm$.003} & 86.08{\tiny$\pm$.004} & \textbf{90.38{\tiny$\pm$.003}} & 81.54{\tiny$\pm$.003} & \textbf{83.38} \\

    \midrule
     
    \rowcolor{gray!20}\cellcolor{white}TabRag$_\text{8B}$ & 72.30{\tiny$\pm$.003} & \textbf{85.61{\tiny$\pm$.001}}  &  84.07{\tiny$\pm$.004}  & 88.15{\tiny$\pm$.003} & 80.01{\tiny$\pm$.003} & 82.03 \\
    
    \rowcolor{gray!20}\cellcolor{white}TabRag$_\text{32B}$ & 72.77{\tiny$\pm$.004} & 85.60{\tiny$\pm$.002}  &  83.58{\tiny$\pm$.002}  & 89.54{\tiny$\pm$.001} & 78.77{\tiny$\pm$.003} & 82.05 \\    
    \bottomrule
    \end{tabular}}
    \end{sc}
    \end{threeparttable}
    \end{normalsize}
    \vspace{-2mm}
\end{table*}

\begin{table*}[!t]
    \centering
    \caption{Leave-one-out ablation study of individual $\text{TabRAG}_{\text{8B}}$ modules on generation accuracy (\%) across five benchmarks. Each row (prefixed with ``$-$'') represents the performance of the pipeline after removing a specific component: Layout (localized segmentation), Self-Gen ICL (autonomous supervision), Structured Representation (hierarchical JSON schema), or Page Overview (global context).}
    \label{tbl:loo-ablation}
    \vspace{-2mm}
    \begin{normalsize}
    \begin{threeparttable}
    \begin{sc}
    \resizebox{\textwidth}{!}{
    \begin{tabular}{lcccccc}
    \toprule
    & \multicolumn{6}{c}{Accuracy (\%) $\uparrow$} \\
    \cmidrule(lr){2-7}
    Models & TAT-DQA & MP-DocVQA & WikiTQ & TableVQA & ComTQA & Average \\
    \midrule

    \rowcolor{gray!20}\cellcolor{white}TabRAG$_\text{8B}$ & \textbf{92.56{\tiny$\pm$.002}} &  \textbf{88.87{\tiny$\pm$.003}} & \textbf{68.49{\tiny$\pm$.003}} & \textbf{70.60{\tiny$\pm$.004}} & \textbf{82.60{\tiny$\pm$.002}} & \textbf{80.62} \\

    -- Layout & 90.97{\tiny$\pm$.002} & 80.15{\tiny$\pm$.003} & 63.60{\tiny$\pm$.003} & 66.60{\tiny$\pm$.002} & 73.30{\tiny$\pm$.002} & 74.92 \\

    -- Self-Gen ICL & 90.12{\tiny$\pm$.001} & 87.41{\tiny$\pm$.002} & 62.04{\tiny$\pm$.002} & 61.20{\tiny$\pm$.000} & 77.40{\tiny$\pm$.001} & 75.63 \\

    -- Structured Rep. + Self-Gen ICL & 87.26{\tiny$\pm$.001} & 87.21{\tiny$\pm$.004} & 62.23{\tiny$\pm$.003} & 61.20{\tiny$\pm$.003} & 70.00{\tiny$\pm$.002} & 73.58 \\

    -- Page Overview & 91.69{\tiny$\pm$.001} & 80.47{\tiny$\pm$.004} & 67.73{\tiny$\pm$.002} & 67.20{\tiny$\pm$.003} & 73.10{\tiny$\pm$.003} & 76.04 \\
    
    \bottomrule
    
    \end{tabular}}
    \end{sc}
    \end{threeparttable}
    \end{normalsize}
    \vspace{-2mm}
\end{table*}

\subsection{Baseline Performance}

\textbf{Generation}. We evaluate TabRAG on downstream generation tasks to isolate the impact of different document parsing methodologies on answer quality. To independently assess generation performance, we operate under an oracle retrieval setting, assuming that the ground-truth document corresponding to each query is provided. This setup ensures that differences in accuracy are attributable to the structural representation of the document rather than retrieval noise. Specifically, we utilize these ground-truth documents to construct the various representations, which then serve as the context for the generative reasoning stage. In an in-context learning \cite{brown2020languagemodelsfewshotlearners} setting, we prompt an LLM (Qwen3-8B \cite{zhang2025qwen3embeddingadvancingtext}) to use the information from the document to answer a question. We observe in Table \ref{tbl:generation} that across all benchmarks, \textsc{TabRAG$_\text{8B}$} and \textsc{TabRAG$_\text{32B}$} exhibit consistent and substantial gains over baselines. By aligning visual, structural, and textual cues before generation, \textsc{TabRAG} enables the language model to reason over documents in a manner that preserves both local tabular relationships and global contextual meaning.

\textbf{Retrieval}. We evaluate \textsc{TabRAG}'s ability to retrieve the most relevant documents corresponding to each query. The datasets TAT-DQA, TableVQA, and ComTQA consist of separate individual document pages. Thus, there is no designated ``group'' of coherent pages available for retrieval experiments. To solve this, we randomly sample the pages into folders of 25 pages each and conduct retrieval within these folders. Given a query, the model searches over the indexed document representations generated by different parsing methods and ranks them based on semantic similarity. We report the mean reciprocal rank at 10 (MRR@10) as the primary evaluation metric. As shown in Table~\ref{tbl:reject_answering}, \textsc{TabRAG} achieves competitive retrieval accuracy compared to existing baselines. We emphasize that our method is primarily designed for improved document understanding and question answering, which enhances generation performance. Therefore, an optimal method of deploying \textsc{TabRAG} is to combine it with existing state-of-the-art retrievers and leverage the corresponding \textsc{TabRAG} vector stores during generation. We leave this to future work.

\begin{table*}[!t]
    \centering
    \caption{Comparison of generation accuracy (\%) across varying structured representations within the TabRAG framework. We evaluate the performance of our hierarchical JSON schema ($\textsc{TabRAG}_{\text{JSON}}$) against standard markdown formatting ($\textsc{TabRAG}_{\text{MD}}$) and natural language interpretation ($\textsc{TabRAG}_{\text{TEXT}}$).}
    \label{tbl:varying_rep_md_text_json}
    \vspace{-2mm}
    \begin{normalsize}
    \begin{threeparttable}
    \begin{sc}
    \resizebox{\textwidth}{!}{
    \begin{tabular}{lcccccc}
    \toprule
    & \multicolumn{6}{c}{Accuracy (\%) $\uparrow$} \\
    \cmidrule(lr){2-7}
    Models & TAT-DQA & MP-DocVQA & WikiTQ & TableVQA & ComTQA & Average \\
    \midrule
    
    \rowcolor{gray!20}\cellcolor{white}TabRAG$_\text{JSON}$ & \textbf{92.56{\tiny$\pm$.002}} &  \textbf{88.87{\tiny$\pm$.003}} & \textbf{68.49{\tiny$\pm$.003}} & \textbf{70.60{\tiny$\pm$.004}} & \textbf{82.60{\tiny$\pm$.002}} & \textbf{80.62} \\
    
    TabRAG$_{\text{MD}}$ & 90.48{\tiny$\pm$.002} & 88.35{\tiny$\pm$.003} & 62.42{\tiny$\pm$.002} & 62.00{\tiny$\pm$.003} & 75.90{\tiny$\pm$.002} & 75.83 \\

    TabRAG$_{\text{Text}}$ & 90.97{\tiny$\pm$.002} & 86.45{\tiny$\pm$.003} & 62.03{\tiny$\pm$.003} & 68.00{\tiny$\pm$.003} & 78.40{\tiny$\pm$.002} & 77.17 \\
    
    \bottomrule
    
    \end{tabular}}
    \end{sc}
    \end{threeparttable}
    \end{normalsize}
    \vspace{-2mm}
\end{table*}

\begin{figure*}[!t]
    \centering
    \includegraphics[width=\textwidth]{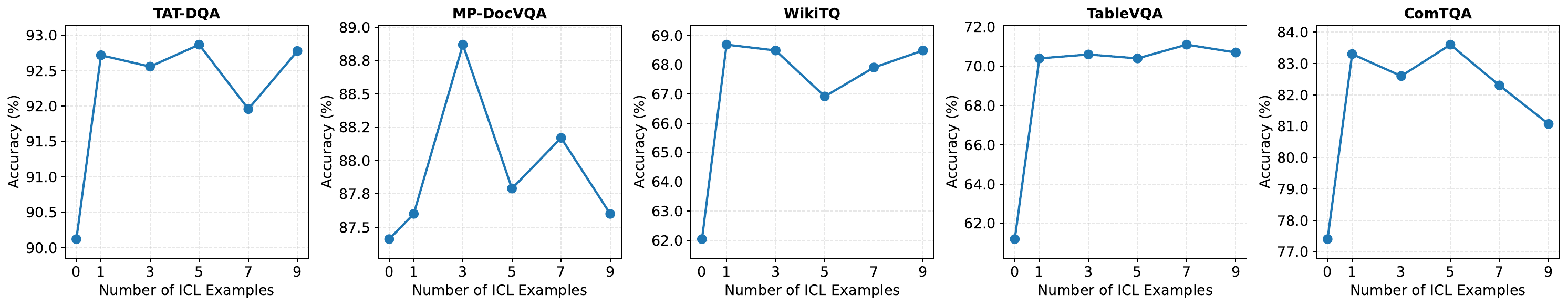}
    \caption{Generation performance in accuracy (\%) of \textsc{TabRAG${_\text{8B}}$} with varying number of self-generated in-context learning examples ($K$). We find three ICL examples ($K=3$) as the sweet spot for guiding the VLM's extraction. Beyond this point, additional demonstrations provide diminishing returns and, in some cases, lead to slight performance degradation, likely due to increased context saturation.}
    \label{fig:num_icl}
    \vspace{-3mm}
\end{figure*}

\subsection{Ablation Studies}

\textbf{TabRAG Modules}. We investigate the impact of the different modules in our \textsc{TabRAG} pipeline. In Table \ref{tbl:loo-ablation}, we conduct an ablation study on the generation performance in which we leave a module out of \textsc{TabRAG}. The modules include layout segmentation, self-generated ICL examples, both the structured representation and the self-generated ICL examples, as well as the page overview. Our results illustrate that each component contributes positively and is crucial to the overall reasoning and extraction capability of the model, as the full \textsc{TabRAG} configuration consistently outperforms all ablated variants across all five datasets.

We provide qualitative insights to support the importance of each component. The core contribution of our method lies in the structured representation that extracts table values for LLM understanding, coupled with self-generated ICLs that aid the VLM in extraction. However, we notice that when parsing structured representations from entire page documents containing a significant amount of information, such as text and complicated tables, the VLM alone is unable to parse the information accurately. It often hallucinates, producing inaccurate values or nonsensical outputs, such as delimiters or empty strings. Layout segmentation helps to solve this issue by breaking pages into focused components. In addition, we notice that without our page overview module, there is a significant decrease in accuracy, especially in the MP-DocVQA and ComTQA datasets. Upon further investigation, we observe that the layout segmentation model occasionally fails to generate reliable bounding boxes, missing key information. Thus, as a safeguard, the page overview ensures that no global context is left out.

\textbf{Structured Representation Design}. While tables are inherently two-dimensional, LLMs require them to be flattened into a one-dimensional representation before they can be indexed. We compare our JSON representation against markdown and text. This is reflected in both the rules of the structured representation and our self-generated ICL. As shown in Table \ref{tbl:varying_rep_md_text_json}, the JSON serialization format (\textsc{TabRAG}$_\text{JSON}$) provides the most robust results. The prompts of this ablation study can be found in Appendix \ref{app:prompts_ablation}. 

The industry standard for parsing tabular documents is to directly serialize or parse the document \cite{llamaindex, langchain}. Therefore, many VLMs \cite{qwen3technicalreport, openai2025gpt52} default to a markdown representation for tables because it provides a linear text format that most closely mimics the visual two-dimensional layout of a table. Another intuitive option is to directly describe the table values using natural language. In Appendix \ref{app:output_ablation-struc-rep}, we perform a case study of these representations and illustrate their outputs in the generation experiment. Overall, we observe that the generation performance degrades as the size of the table scales with the number of columns and rows. The issue stems from the sequential nature of LLM processing, which makes it difficult for the model to maintain the mental link between the data points and the structural anchors required for multi-step reasoning. If we visualize passing in the ``flattened'' \textsc{TabRAG}$_\text{MD}$ table, the values in the right-most and bottom-most columns and rows lose their ability to associate with their corresponding row and column names. The parsing operates in a raster-scan manner, causing the names to be parsed much earlier than the values. Similarly, for a textual representation, we notice that the lack of structure inherently prevents the model from synthesizing information across multiple rows and columns. While the JSON representation explicitly links each value to its metadata, the text and markdown-based generations fail to maintain these relational threads. This is demonstrated in both \textsc{TabRAG}$_\text{MD}$ and \textsc{TabRAG}$_\text{Text}$ generation outputs, where the model incorrectly claims that total assets ``cannot be directly calculated'' and are ``not explicitly provided,'' despite the data being physically present in the source. Ultimately, the lack of ``structure'' leads the model to perceive the table as a collection of disjointed components rather than a cohesive data set, preventing it from performing basic arithmetic or identifying summary figures.

\textbf{Number of Self-Gen ICL Demonstrations}. While increasing the number of demonstrations in the prompt generally improves the model’s task alignment, the increase in prompt length introduces a trade-off against the model's capacity to reason over longer contexts \cite{liu2023lostmiddlelanguagemodels}. We analyze the effect of the number of self-generated ICL demonstrations on our generation performance. As observed in Figure \ref{fig:num_icl}, performance improves sharply when moving from zero to a small number of demonstrations, with three ICL examples consistently yielding stable gains across most datasets. Beyond this point, additional demonstrations provide diminishing returns and, in some cases, lead to slight performance degradation, likely due to increased context saturation.
\section{Related Works}

\textbf{Document Understanding}. Document AI has transitioned from raw OCR toward unified multimodal architectures like LayoutLMv3 \cite{huang2022layoutlmv3pretrainingdocumentai} and DocFormer \cite{appalaraju2021docformerendtoendtransformerdocument}. While these models effectively integrated text and layout, their encoder-only nature was optimized for discriminative labeling rather than the reasoning required for complex QA. Conversely, generative frameworks such as UDOP \cite{tang2023unifyingvisiontextlayout} and Nougat \cite{blecher2023nougatneuralopticalunderstanding} relied on linearized sequences to represent document content. In tabular tasks, this sequential approach failed to maintain the relational associations between headers and data points as sequence lengths grew. TabRAG addresses these limitations by treating tabular QA as a localized task, utilizing hierarchical representations to explicitly preserve the 2D semantics that holistic or sequential frameworks are not architecturally biased to recover.

\textbf{RAG for Visually Rich Documents}. In the context of visually rich documents, standard RAG often failed to bridge the ``semantic gap" created when text-only extraction discarded essential layout cues. Recent pipelines like VisRAG \cite{yu2025visragvisionbasedretrievalaugmentedgeneration} and ColPali \cite{faysse2025colpaliefficientdocumentretrieval} addressed this by utilizing vision-language models to produce document embeddings that incorporated visual and spatial features directly. While these vision-based retrieval methods improved relevance for retrieval, they often lacked the granularity to preserve specific cell-to-header associations in dense tables for generation. To maintain relational context, GraphRAG \cite{edge2025localglobalgraphrag} and StructRAG \cite{li2024structragboostingknowledgeintensive} utilized LLMs to reconstruct information into knowledge graphs or dynamic formats at inference time. However, these methods relied on probabilistic inference from text chunks, which often lack the spatial-logical cues required to recover the precise 2D hierarchy of a tabular document. In contrast, \textsc{TabRAG} deterministically maps the physical 2D hierarchy of a table into a hierarchical structural rationale, allowing for the preservation of explicit spatial-semantic information especially from dense, cell-heavy PDFs.

\textbf{Table Understanding}.
Table understanding research has primarily focused on discrete reasoning over data that is already structured. Chain-of-Table \cite{wang2024chainoftableevolvingtablesreasoning} introduced a dynamic framework for iteratively transforming tables to mimic human thought, while TableRAG \cite{yu2025tableragretrievalaugmentedgeneration} utilized SQL-based querying for explicit data retrieval. However, these methods assume a ``clean" input, typically in HTML, CSV, or SQL formats, thereby bypassing the fundamental challenge of interpreting raw, unstructured document (PDF) pixels. While UniTable \cite{peng2024unitableunifiedframeworktable} unified structure recognition and content extraction into a pixel-to-sequence objective, its focus on table recognition and reliance on linear decoding remains a bottleneck for preserving relational dependencies in large-scale tables. To the best of our knowledge, \textsc{TabRAG} is the only framework that offers a complete end-to-end pipeline that specifically targets tabular document QA within RAG frameworks. Rather than assuming a pre-parsed schema, \textsc{TabRAG} operates directly on document images, converting them into a hierarchical structural rationale. This ensures that the reasoning engine receives the explicit two-dimensional context required to bridge the gap between raw document pixels and high-level discrete reasoning.

\textbf{Agentic Rationale Generation}. Recent approaches in document parsing have moved beyond simple extraction, incorporating agentic rationale generation \cite{shinn2023reflexionlanguageagentsverbal, yuksekgonul2024textgradautomaticdifferentiationtext, xiao2025verbalizedmachinelearningrevisiting} to improve reliability. In the text domain, frameworks like Self-RAG introduced self-reflection tokens to dynamically critique retrieval quality \cite{asai2023selfraglearningretrievegenerate}, while InstructRAG leveraged self-synthesized rationales to explicitly filter noise from retrieved contexts \cite{wei2025instructraginstructingretrievalaugmentedgeneration}. This reasoning-centric paradigm is now influencing visual document understanding, where methods such as Rationale Distillation \cite{jia2025bridgingrelevancereasoningrationale} train models to articulate intermediate reasoning steps for complex visual tasks. In this work, \textsc{TabRAG} adopts this philosophy by utilizing VLMs to generate structured JSON representations with agentic self-generated ICLs, ensuring that extracted values are explicitly grounded in their spatial and semantic layout.

\section{Conclusion}
In this work, we present \textsc{TabRAG}, a retrieval augmented generation framework for improving tabular document question answering via structured representations. We first segment the documents into various layout components. Specifically for table components, we design a prompt that extracts a representation to effectively capture structural and semantic information from tabular documents. We couple our extraction process with a self-generated in-context learning module to provide information regarding the complexities of the table's design as well as domain-specific knowledge. We conducted extensive experiments and comprehensive ablations to evaluate \textsc{TabRAG}. Our results on multiple tabular-document question answering datasets demonstrate \textsc{TabRAG}'s prowess in generation tasks while remaining competitive in retrieval. \textsc{TabRAG}, coupled with state-of-the-art retrievers, can help bridge the gap between table understanding and document-level comprehension.

\newpage


    




    
    

\section*{Impact Statement}
The goal of this work is to make AI-driven tabular document parsing and analysis more reliable by improving reasoning on complex tables. While standard AI often ``hallucinates'' when looking at dense spreadsheets or financial reports, TabRAG focuses on turning those visuals into a clear, structured format that the model can actually reason through.

The broader implications of this approach include:
Higher Accuracy in Critical Fields: In areas like finance, healthcare, or scientific research, misreading a single cell in a table can have major consequences. By grounding AI in the actual structure of a document, we can reduce the risk of "hallucinations" where the AI makes up a number because it couldn't follow the row or column or misinterprets the table's fundamental structure.

Making Data More Accessible: Many complex reports and documents are currently difficult for automated systems to read and parse effectively without human intervention. This technology helps bridge that gap, making it easier for everyone from individual tax-filers to large organizations to get fast, accurate answers from their own data.

Better Transparency and Interpretability: Our method employs a step-by-step parsing process, converting visuals into JSON representations, making it easier for users to verify the origin of an answer. This makes the entire system more predictable and trustworthy compared to "black box" models that process a whole page at once.




\bibliography{biblio}
\bibliographystyle{icml2026}

\newpage
\appendix
\onecolumn

\begin{center}
\Large
\textbf{Appendix}
\end{center}

\etocdepthtag.toc{mtappendix}
\etocsettagdepth{mtchapter}{none}
\etocsettagdepth{mtappendix}{subsection}
{\small \tableofcontents}

\newpage
\section{Implementation}
\label{app:implementation}

\subsection{Hardware Details}
\label{app:hardware}

The following delineates the foundation of our experiments:
\begin{itemize}
    \item Codebase: Python \& PyTorch
    \item CPU: AMD EPYC 7443P
    \item GPU: NVIDIA A100 80GB PCIe / NVIDIA A6000 48GB
\end{itemize}

\subsection{Model Configuration}
\label{app:model_config}

The model configurations used are shown in Table \ref{tbl:model-config}. The hyperparameters for the remaining baselines \cite{pymupdf, pytesseract} are their default configurations. Qwen3-VL-8B-Instruct (\textsc{TabRAG$_{\text{8B}}$}) and Qwen3-VL-32B-Instruct (\textsc{TabRAG$_{\text{32B}}$}) are used as the vision-language models for \textsc{TabRAG}, while Qwen3-8B (No Think) is used for evaluating generation performance.

\begin{table}[htbp] 
    \centering
    \caption{Model Configurations.} 
    \label{tbl:model-config}
    \footnotesize
    \begin{threeparttable}
    \resizebox{0.7\columnwidth}{!}{
    \begin{tabular}{lcc}
        \toprule
        \textbf{Model} & \textbf{Temperature} & \textbf{Context Window} \\
        \midrule 
        Qwen3-VL-8B-Instruct & $1.0$ & $32768$ \\
        Qwen3-VL-32B-Instruct & $1.0$ & $32768$ \\
        Qwen3-8B (No Think) & $1.0$ & $32768$ \\
        \bottomrule 
    \end{tabular}
    }        
  \end{threeparttable}
\end{table}


\newpage
\section{Algorithms}
\label{app:algorithm}

\subsection{TabRAG}
\begin{algorithm*}[!ht]
    \small
    \setstretch{1}
    \caption{\textsc{TabRAG} Framework}
    \label{alg:tabrag}
    \begin{algorithmic}[1]

        \STATE \textbf{Input:} Dataset $\mathcal{D} = \{I_j\}_{j=1}^{N}$, Query $q$, No. of Documents to Retrieve $k$
        \STATE \textbf{Output:} Response $y$

        \STATE $\mathcal{S} \leftarrow \textsc{SelfGenICL}(\mathcal{D})$
        \hfill \textcolor{gray}{\Comment{self-generated ICL module; see Alg.~\ref{alg:icl_sampling}}}

        \STATE $\Omega \leftarrow \emptyset$
        \hfill \textcolor{gray}{\Comment{initialize vector store}}

        \FORALL{$I \in \mathcal{D}$}

            \STATE $\mathcal{R} \leftarrow \emptyset$
            \hfill \textcolor{gray}{\Comment{parsed representations}}

            \STATE $\Pi \leftarrow f_{\text{layout}}(I)$
            \hfill \textcolor{gray}{\Comment{where $\Pi = \{\pi_i\}_{i=1}^{M}$}} 

            \FORALL{$\pi_i = (b_i, x_i, c_i) \in \Pi$}
            \hfill \textcolor{gray}{\Comment{$\pi_i = (b_i, x_i, c_i)$ = (bbox, cropped img, component type)}} 

                \IF{$c_i == \texttt{table}$}
                    \STATE $r_i \leftarrow f_{\text{VLM}}(x_i, p_{c_i}, \mathcal{S})$
                    \hfill \textcolor{gray}{\Comment{parse table with ICL supervision}}
                \ELSE
                    \STATE $r_i \leftarrow f_{\text{VLM}}(x_i, p_{c_i})$
                    \hfill \textcolor{gray}{\Comment{parse non-table component}}
                \ENDIF

                \STATE $\mathcal{R} \leftarrow \mathcal{R} \cup \{r_i\}$

            \ENDFOR

            \STATE $\mathcal{R} \leftarrow \mathcal{R} \cup \{f_{\text{VLM}}(I, p_{page})\}$ \hfill \textcolor{gray}{\Comment{page-level overview}}

            \STATE $\Omega \leftarrow \Omega \cup \{(f_{\text{emb}}(\mathcal{R}), \mathcal{R})\}$
            \hfill \textcolor{gray}{\Comment{embed document and add to vector store}}

        \ENDFOR

        \STATE $\mathcal{R}_q \leftarrow \textsc{RetrieveTopK}(q, \Omega, k)$
        \hfill \textcolor{gray}{\Comment{retrieve top-$k$ documents}}

        \STATE $y \leftarrow f_{\text{LLM}}(q, \mathcal{R}_q)$
        \hfill \textcolor{gray}{\Comment{generate response}}

    \end{algorithmic}
\end{algorithm*}

\newpage
\section{Experiments}
\label{app:expts}

\subsection{Datasets}
\label{app:datasets}

Prior to constructing the ragstores, the datasets undergo preprocessing. Each file in the dataset can contain multiple pages; thus, for consistency, we ensure that each file is separated into individual files containing a single page each. For methods that require image files as inputs, we convert data modalities such as PDFs and HTML files into high-resolution PNG files.

\begin{table}[htbp] 
    \centering
    \caption{Dataset Overview.} 
    \label{tbl:exp-dataset}
    \footnotesize
    \begin{threeparttable}
    \resizebox{0.9\columnwidth}{!}{
    \begin{tabular}{lccc}
        \toprule
        \textbf{Dataset} & \textbf{Document Type} & \textbf{\# Pages} & \textbf{\# QA Pairs} \\
        \midrule 
        TAT-DQA \cite{tatdqaZhu_2022} & Financial & $312$ & $1640$ \\
        MP-DocVQA \cite{mpdocvqatito2023hierarchicalmultimodaltransformersmultipage} & Multi-domain & $500$ & $515$ \\
        WikiTQ \cite{wikitqpasupat2015compositionalsemanticparsingsemistructured} & Web Tables & $243$ & $511$ \\
        TableVQA-Bench \cite{kim2024tablevqabench} & Multi-domain & $427$ & $500$ \\
        ComTQA \cite{zhao2024tabpedia_comtqa} & Multi-domain & $500$ & $1000$ \\
        \bottomrule 
    \end{tabular}
    }        
  \end{threeparttable}
\end{table}

In the following section, we provide a description and visualizations of the datasets.

\newpage
\textbf{TAT-DQA} \cite{tatdqaZhu_2022}. TAT-DQA comprises PDF document pages that include both semi-structured tables and unstructured textual content, each paired with corresponding question–answer sets. The documents are drawn from financial reports and feature a strong presence of numerical information. The dataset consists of separated individual PDF pages thus, there is no designated ``group'' of pages available for retrieval experiments. Hence, we first sort the document pages then split these pages into groups of 25. This yields 13 folders which we use to construct ragstores for retrieval.
\begin{figure}[t]
    \centering
    \includegraphics[width=0.8\textwidth]{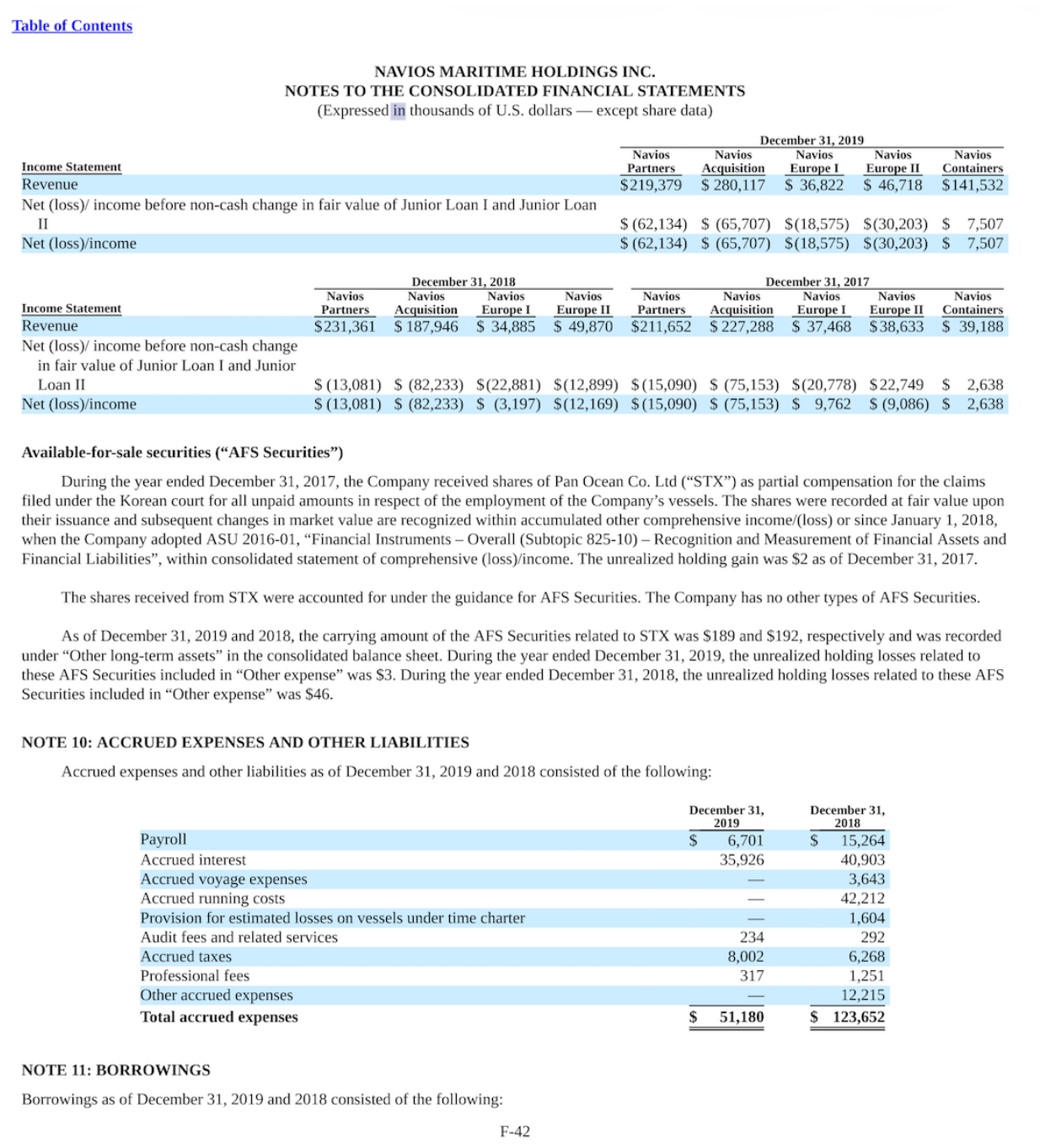}
    \caption{An example page from the TAT-DQA dataset illustrating financial tables interleaved with descriptive text and numerical values.}
    \label{fig:tatdqa_example}
    \vspace{-3mm}
\end{figure}

\clearpage
\textbf{MP-DocVQA} \cite{mpdocvqatito2023hierarchicalmultimodaltransformersmultipage}. Document Visual Question Answering (DocVQA) datasets answer questions from document images. MP-DocVQA extends DocVQA \cite{mathew2021docvqa} to the multi-page scenario where DocVQA contains images such as typed or handwritten text, layout, diagrams, figures, tabular structures, and signatures. The dataset consists a total of 46K questions, 6K documents, and 48K pages (Images). We filter the dataset by selecting documents that contains tables (via layout detection), followed by documents with a high number of questions to pages ratio.
\begin{figure}[t]
    \centering
    \includegraphics[width=0.7\textwidth]
    {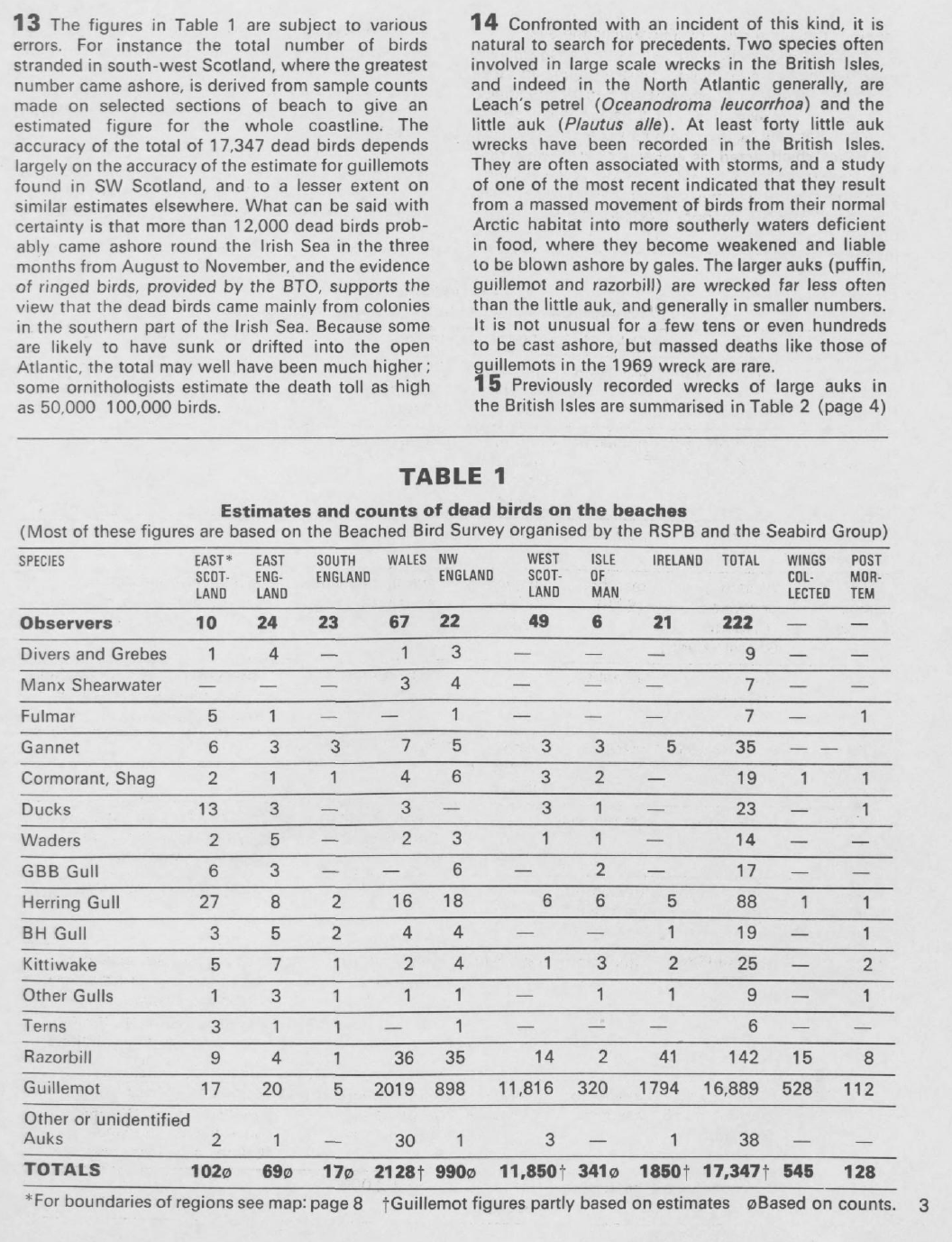}
    \caption{A sample document image from MP-DocVQA showing a scanned document page with dense tables and multi-column text.}
    \label{fig:mpdocvqa_example}
    \vspace{-3mm}
\end{figure}

\clearpage
\textbf{WikiTableQuestions} \cite{wikitqpasupat2015compositionalsemanticparsingsemistructured}. WikiTableQuestions aim to answer complex questions about semi-structured HTML tables using question-answer pairs as supervision. The dataset consists of question-answer pairs corresponding to an accompanying table. The questions include a variety of operations such as comparison, superlatives, aggregation, and arithmetic.
\begin{figure}[t]
    \centering
    \includegraphics[width=0.55\textwidth]{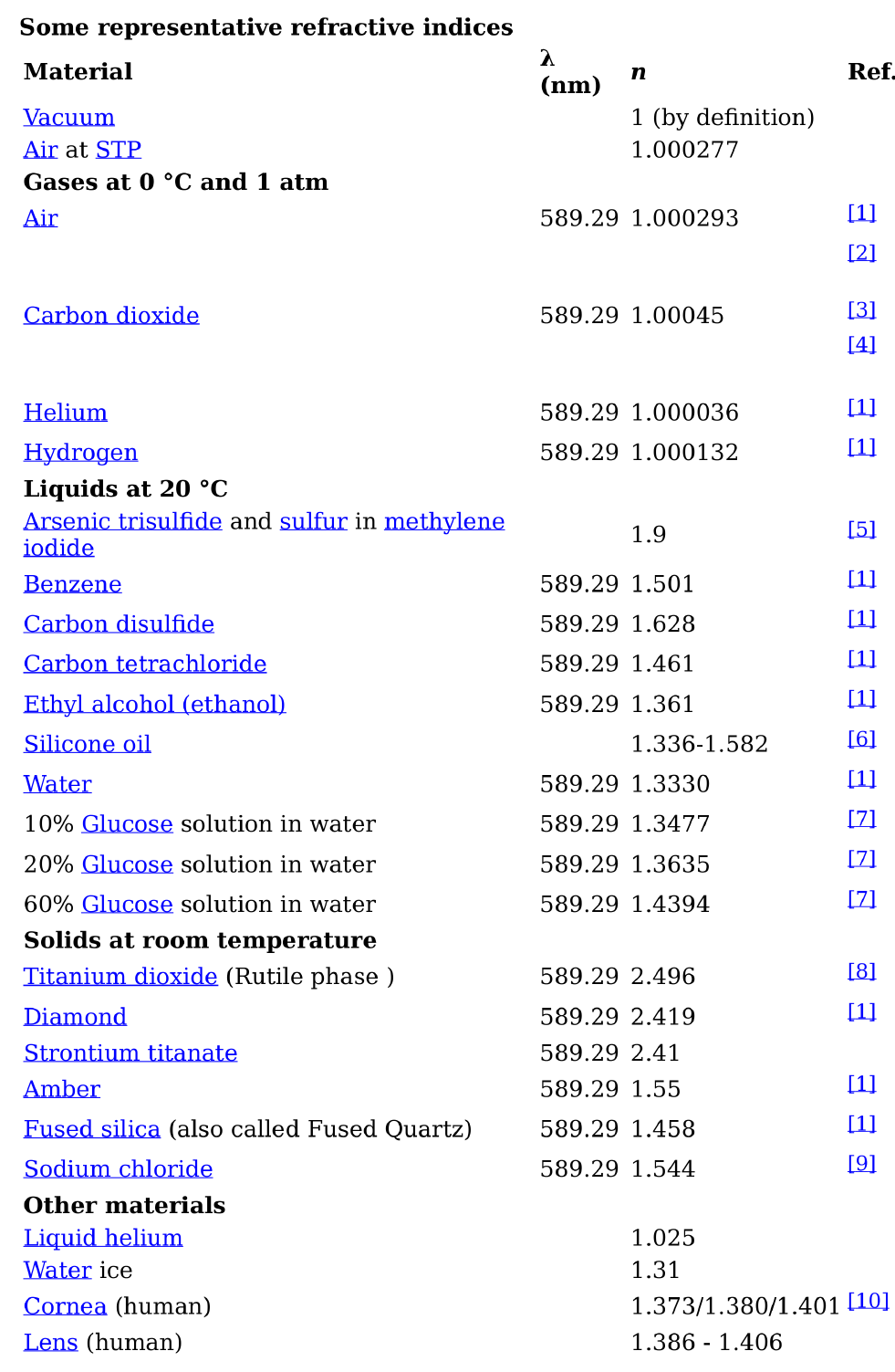}
    \caption{A rendered web table from the WikiTableQuestions dataset illustrating semi-structured tabular data with numerical and categorical entries.}
    \label{fig:wikitq_example}
    \vspace{-3mm}
\end{figure}

\clearpage
\textbf{TableVQA-Bench} \cite{kim2024tablevqabench}.
TableVQA-Bench is a unified benchmark that aggregates multiple table-centric datasets, including VWTQ, VWTQ-Syn, VTabFact, and FinTabNetQA, covering diverse domains such as web tables, fact verification, and financial reports. To manage the size of the dataset in our experiments (number of documents), we use the FinTabNetQA and VTabFact subsets and treat each table image as an independent document. The dataset consists of separated individual PNG pages. Thus, there is no designated “group” of pages available for retrieval experiments. Hence, we first sort the document pages then split these pages into groups of 25. This yields 18 folders which we use to construct ragstores for retrieval.
\begin{figure}[t]
    \centering
    \includegraphics[width=\textwidth]{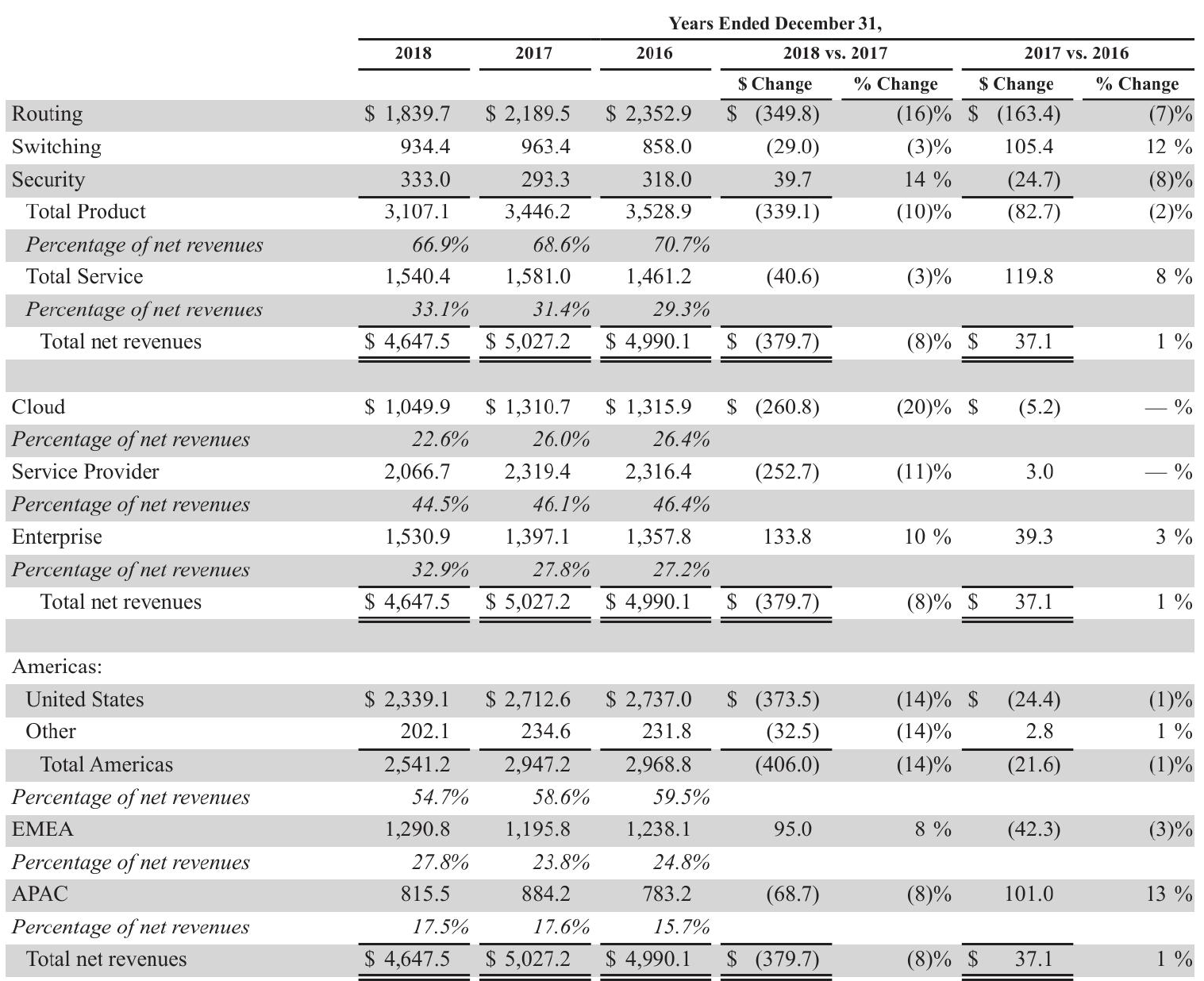}
    \caption{An example table from the TableVQA-Bench dataset with semi-structured layouts and mixed numerical and categorical content.}
    \label{fig:tablevqa_example}
    \vspace{-3mm}
\end{figure}

\clearpage
\textbf{ComTQA} \cite{zhao2024tabpedia_comtqa}.
ComTQA is a visual table question answering dataset composed of table images paired with question–answer annotations. The dataset aggregates tables from PubTables-1M and FinTabNet, spanning scientific and financial domains. The dataset consists of separated individual PNG pages. Thus, there is no designated “group” of pages available for retrieval experiments. Hence, we first sort the document pages then split these pages into groups of 25. This yields 20 folders which we use to construct ragstores for retrieval.
\begin{figure}[t]
    \centering
    \includegraphics[width=\textwidth]{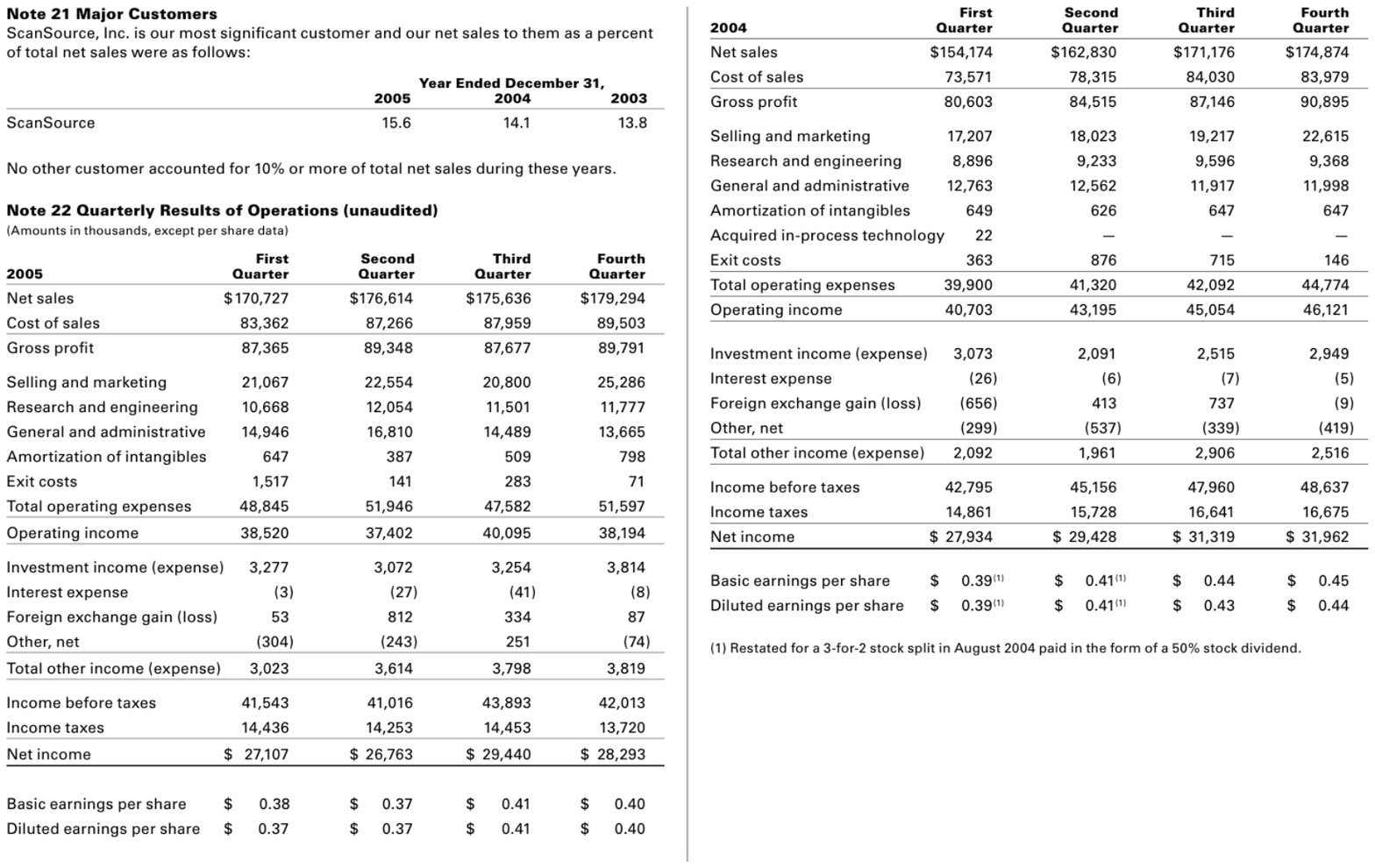}
    \caption{An example table from the ComTQA dataset with semi-structured layouts and mixed data types.}
    \label{fig:comtqa_example}
    \vspace{-3mm}
\end{figure}

\clearpage
\subsection{Baselines}
\label{app:baselines}

\textsc{TabRAG}'s performance is evaluated in comparison to popular parsing methods. This includes PyMuPDF \cite{pymupdf}, PyTesseract \cite{pytesseract} and multiple vision language models drawn from the DeepSeek \cite{wei2025deepseekocr}, Gemma \cite{gemma2025gemma3}, GPT \cite{openai2025gpt52} and Qwen \cite{qwen3technicalreport} model families.

\textbf{PyMuPDF} \cite{pymupdf}. PyMuPDF is a Python library for data extraction and analysis of PDF documents. It interfaces with the MuPDF rendering engine, which parses the PDF’s internal object structure-decoding vector drawing commands, text placements, and embedded resources-to reconstruct the page layout.

\textbf{PyTesseract} \cite{pytesseract}. PyTesseract is a Python wrapper for Google’s Tesseract OCR engine \cite{10.5555/1288165.1288167}, used for extracting text from images and scanned documents. It performs a raster-based analysis of the input image-segmenting it into connected components, detecting character shapes, and matching them against trained language models-to recognize and reconstruct textual content from pixel patterns.

\textbf{DeepSeek-OCR} \cite{wei2025deepseekocr}. DeepSeek-OCR is an OCR-oriented vision system that explores compressing long textual contexts via optical 2D mapping of document content. It consists of a high-resolution visual encoder (DeepEncoder) that produces a set of vision tokens and a decoder (DeepSeek3B-MoE-A570M) that generates recognized text from these representations. By explicitly controlling the vision-token budget and compression ratio, DeepSeek-OCR enables efficient text decoding under constrained visual-token settings, making it suitable as an OCR-based document parsing baseline.

\textbf{Gemma-3-27B-IT} \cite{gemma2025gemma3}. Gemma-3-27B-IT is a large instruction-tuned multimodal language model from the Gemma 3 family developed by Google DeepMind. It integrates visual understanding and long-context processing within a unified transformer architecture, with architectural modifications to mitigate KV-cache growth. Trained via distillation and post-training refinement, the model supports structured instruction following and multimodal reasoning over document images, tables, and text.


\textbf{GPT-5.2} \cite{openai2025gpt52}.
GPT-5.2 is a state-of-the-art proprietary multimodal large language model developed by OpenAI for professional knowledge work and complex reasoning tasks. The model integrates advanced capabilities in natural language understanding, long-context processing, and tool usage within a unified architecture. GPT-5.2 demonstrates strong performance on multi-step reasoning, document and chart understanding and code generation that require planning and execution over long horizons.

\textbf{Qwen3-VL-32B/8B-Instruct} \cite{qwen3technicalreport}. 
Qwen3-VL-32B/8B-Instruct is a large-scale vision-language model developed by Alibaba Cloud. It integrates visual and textual understanding through a unified multimodal transformer architecture capable of processing both images and natural language. The model is trained on large-scale multimodal corpora, enabling it to interpret complex document layouts, tables, and figures while aligning visual cues with textual semantics. Its instruction-tuned variant enhances performance on document understanding and question answering tasks by following structured prompts and adapting to diverse multimodal reasoning scenarios.

\subsection{Benchmarks}
\label{app:benchmarks}

\textbf{Accuracy} \cite{yu2025visragvisionbasedretrievalaugmentedgeneration, faysse2025colpaliefficientdocumentretrieval}. Accuracy is computed by evaluating the model’s ability to produce correct answers to the ground-truth questions. For each document, the corresponding docstore and question are passed to the language model to generate an answer. The generated response is normalized and compared against the normalized ground-truth answers. A prediction is counted as correct if all ground-truth answers appear in the model’s response. The final accuracy is calculated as the ratio of correctly answered questions to the total number of evaluated questions.


\textbf{Mean Reciprocal Rank} \cite{yu2025visragvisionbasedretrievalaugmentedgeneration}. 
Mean Reciprocal Rank (MRR@10) evaluates the ranking quality of the retrieved document components relative to the ground-truth evidence. 
For each question, the corresponding document’s vector store is loaded, and the question embedding is used to retrieve the top-10 most relevant components. 
The reciprocal rank is computed based on the first retrieved document containing the ground-truth evidence. 
The final MRR@10 score is obtained by averaging the reciprocal ranks across all evaluated questions, as defined by
\begin{equation}
    \text{MRR} = \frac{1}{|Q|} \sum_{i=1}^{|Q|} \frac{1}{\text{rank}_i},
\end{equation}
where $|Q|$ is the total number of questions and $\text{rank}_i$ denotes the rank position of the first relevant retrieved component for the $i$-th query.

\newpage
\section{Prompts}
\label{app:prompts}

\subsection{TabRAG: Self-Generated In-Context Learning}
\label{app:prompts_selfgen-icl}

In this section, we detail the prompts used to construct our Self-Generated In-Context Learning (ICL) examples. We begin by creating a representative markdown string to textually represent the target table. To achieve this, we utilize the prompt below, which includes several rudimentary ICL examples that provide concrete demonstrations of the extraction rules.

\begin{ttcolorbox}[TabRAG Self-Generated ICL: Markdown]
\begin{minipage}{\linewidth}
\begin{lstlisting}[]
You are a precise information parsing and extraction engine. Output ONLY
A markdown table of the provided image

Task: Generate a markdown table with the correct structure, with every single cell accounted for.

Rules:
1. Output ONLY a valid Markdown table. Do not include any other text.
2. Preserve the exact number of rows and columns visible in the image.
3. Preserve unit information in headers if present (e.g., "$ in millions").
4. Transcribe all cell contents as faithfully as possible without inference, normalization, or computation.
5. Repeat values for merged cells so that every row has the same number of columns.
6. If a cell is visually empty, leave it empty.
7. Ignore captions, footnotes, and non-table elements outside the table.


---

**Example Output 1: Two-level header**

| ($ in millions) | 2024             | 2023             |
|-----------------|------------------|------------------|
|                 | Revenue | Profit | Revenue | Profit |
| Sales           | 1,234   | 400    | 1,200   | 350    |
| Net Income      | (56)    | 80     | -40     | 70     |

---

**Example Output 2: Three-level header**

| ($ in thousands) | 2024                                 | 2023                                |
|------------------|--------------------------------------|-------------------------------------|
|                  | Q1                | Q2               | Q1               | Q2               |
|                  | Revenue | Profit  | Revenue | Profit | Revenue | Profit | Revenue | Profit |
| Product A        | 500     | 120     | 600     | 150    | 450     | 100    | 550     | 140    |
| Product B        | (50)    | 80      | (30)    | 100    | -20     | 60     | 10      | 90     |

---

**Example Output 3: Mixed 1-row, 2-row, and 3-row headers**

| Category              | 2024                                | 2023             | Growth % | Notes    |
|-----------------------|-------------------------------------|------------------|----------|----------|
|                       | Q1               | Q2               | Revenue | Profit |          |          |
|                       | Revenue | Profit | Revenue | Profit |         |        |          |          |
| Sales (Units)         | 1,000   | 300    | 900     | 250    | 1,700   | 550    | 12%      | N/A      |
| Cost ($ in thousands) | (200)   | (50)   | -180    | -40    | (380)   | (90)   | N/A      | Adjusted |

---

Now, generate the markdown table ONLY, do not include any other irrelevant text.
\end{lstlisting}
\end{minipage}
\end{ttcolorbox}

Additionally, we use the VLM to generate in-context examples for our structured JSON representation. This process converts visual tables into a standardized JSON format that maps headers to cell values. Including these exemplars provides the model with the structural grounding to accurately parse complex layouts, such as multi-level headers and merged cells. The prompt is shown below.

\begin{ttcolorbox}[TabRAG Self-Generated ICL: JSON]
\begin{minipage}{\linewidth}
\begin{lstlisting}[]
You are a precise information parsing and extraction engine. Output ONLY
A JSON array of objects, each with: {"row": <string>, "column": <string>, "value": <string|null>, "units": <string|null>}.

Task: Then Extract every visible cell in the attached table image into JSON objects.

Each table cell must be represented as:
{
"row": string,        // the row label (e.g. "Revenue", "2024", "Row 1" if unnamed)
"column": string,     // column header text; if multi-level, join levels with " -> "
"value": string|null,  // exact text as seen in the table (keep symbols and brackets)
"units": string|null // units if present in header (e.g., "$ in millions"), otherwise null
}

\end{lstlisting}
\end{minipage}
\end{ttcolorbox}

After generating both the markdown and JSON representations using the prompts above, we merge them to form the final input-output pairs for in-context learning. In this setup, the markdown string serves as the input context, while the JSON object provides the target structured output. This pairing teaches the model to map visual layouts directly to our ``row -\> column -\> value'' schema.

\begin{ttcolorbox}[TabRAG Self-Generated ICL: JSON + Markdown]
\begin{minipage}{\linewidth}
\begin{lstlisting}[]
**INPUT**:
{Markdown Representation}

**OUTPUT**:
{JSON Representation}
\end{lstlisting}
\end{minipage}
\end{ttcolorbox}

\subsection{TabRAG}
\label{app:prompts_tabrag}

\textbf{Table Component Prompt}.
Following layout segmentation and region classification, \textsc{TabRag} applies component-specific prompt templates conditioned on the semantic type of each detected region. Specifically, for tables, we couple our structured representation prompt along with our self-generated ICL examples.



\begin{ttcolorbox}[TabRAG: Table]
\label{fig:vlm_prompt}
\begin{minipage}{\linewidth}
\begin{lstlisting}[]
You are a precise information extraction engine. Output ONLY a JSON array of objects, each with:
{"row": <string>, "column": <string>, "value": <string|null>, "units": <string|null>}.
No markdown, explanations, or text before/after the JSON.

Task: Extract every visible cell in the attached table image into JSON triples.

Each table cell must be represented as:
{
"row": string,        // the row label (e.g. "Revenue", "2024", "Row 1" if unnamed)
"column": string,     // column header text; if multi-level, join levels with " -> "
"value": string|null, // exact text as seen in the table (keep symbols and brackets)
"units": string|null, // units if present in header (e.g., "$ in millions"), otherwise null
}

Rules:
- Output ONLY a JSON array: [ {row, column, value, units}, ... ].
- Order: top-to-bottom, left-to-right.
- Preserve all text formatting exactly as shown:
- Keep parentheses, minus signs, commas, currency symbols, and percent signs.
- Do NOT normalize numbers or remove punctuation.
- Multi-line text: join with a single space.
- Multi-level headers: join with " -> " (e.g. "2024 -> Revenue").
- If a row header spans multiple rows, repeat its label for each affected row.
- If units are present, include them in the output (e.g., "$ in millions"), otherwise use null.
- Use null only for empty or blank cells.

---

{``Self-Generated ICL''}

Now, extract all visible cells from the attached table image and output only the JSON array of {row, column, value, units} objects 
using the " -> " separator for multi-level headers, keeping all cell values exactly as written in the table. 
ENSURING THAT ALL EXTRACTED VALUES ARE ACCURATE IS THE MOST IMPORTANT! DO NOT OUTPUT ANYTHING ELSE.
\end{lstlisting}
\end{minipage}
\end{ttcolorbox}

\newpage
Here's an example.

\begin{figure}[!ht]
    \centering
    \includegraphics[width=0.95\textwidth]{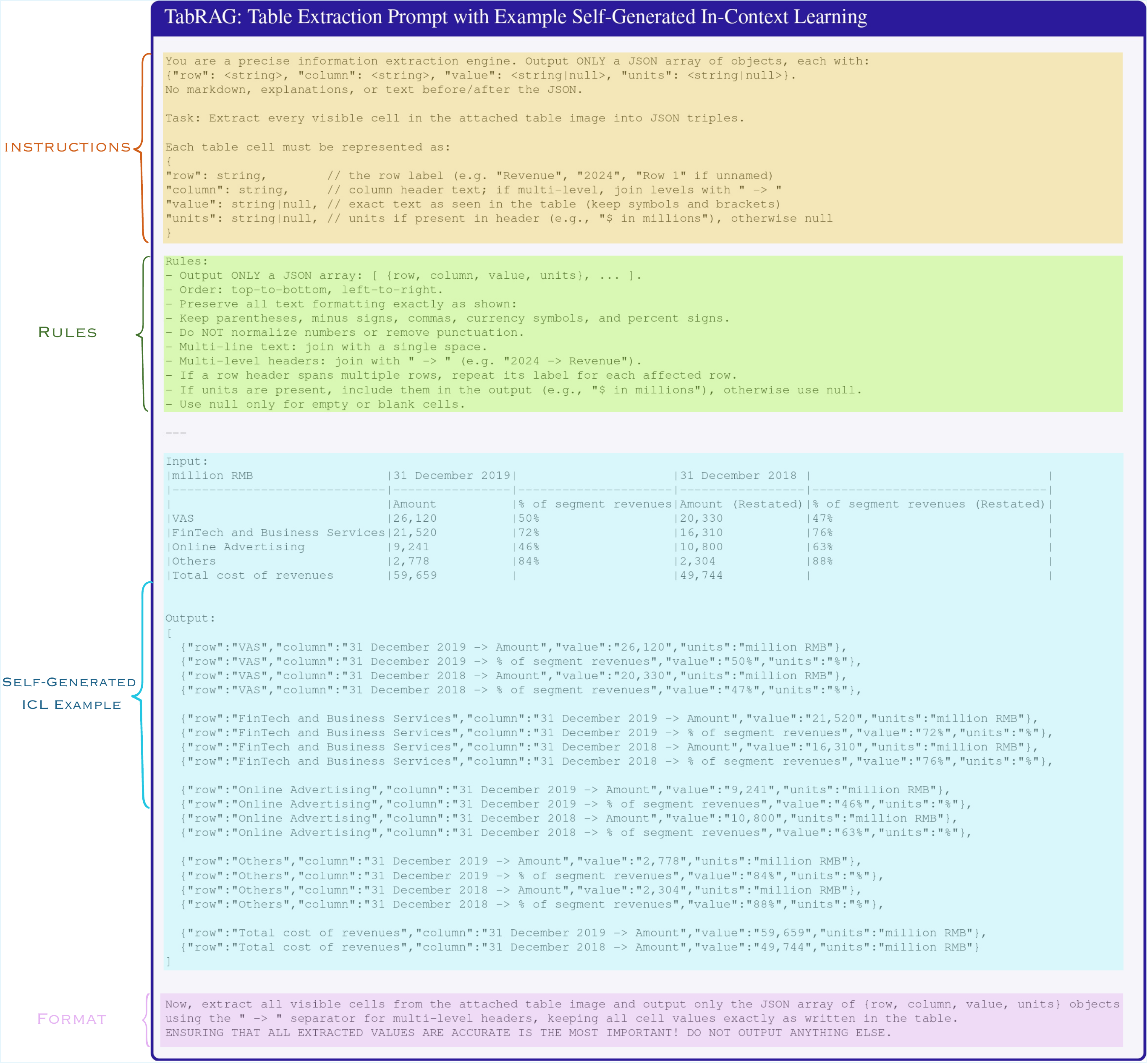}
    \caption{Sample \textsc{TabRAG} prompt with one self-generated ICL example.}
    \label{fig:tabrag_prompt}
    \vspace{-3mm}
\end{figure}

\clearpage
\textbf{Other Component Prompts}. TabRAG also applies VLM prompts to summarize and parse other non-table components. These prompts are shown below.

\begin{ttcolorbox}[TabRAG: Text]
\begin{minipage}{\linewidth}
\begin{lstlisting}[]
Please extract and output the **visible text** in the image exactly **as it appears**, without rephrasing, summarizing, or skipping 
any content. Preserve original formatting such as line breaks, punctuation, and capitalization. This includes any small footnotes 
or embedded labels. DO NOT OUTPUT ANYTHING ELSE!
\end{lstlisting}
\end{minipage}
\end{ttcolorbox}

\begin{ttcolorbox}[TabRAG: Title]
\begin{minipage}{\linewidth}
\begin{lstlisting}[]
Please extract and output the **title text** from the image exactly **as displayed**, preserving capitalization and formatting. Do 
not interpret or rewrite. Output the title as it appears visually. DO NOT OUTPUT ANYTHING ELSE!
\end{lstlisting}
\end{minipage}
\end{ttcolorbox}

\begin{ttcolorbox}[TabRAG: Figure]
\begin{minipage}{\linewidth}
\begin{lstlisting}[]
Please interpret the figure and describe it in detail. Your output should include:
1. Descriptions of individual data points if visible,
2. Descriptions of trend lines, axes, and labels,
3. Explanations of any color or shape encodings, and
4. Any other notable features (e.g., anomalies, clustering, outliers).
Be precise and avoid speculation. Ensure your interpretation **accurately matches the figure** and 
corresponds to what is visually present. DO NOT OUTPUT ANYTHING ELSE!
\end{lstlisting}
\end{minipage}
\end{ttcolorbox}

\textbf{Page Overview Prompt}. We use the following prompt to generate a page overview, which supplements the component-wise representations by providing cross-component context.
\begin{ttcolorbox}[TabRAG: Page]
\begin{minipage}{\linewidth}
\begin{lstlisting}[]
Please parse everything in the attached image and output the parsed contents only without anything else.
\end{lstlisting}
\end{minipage}
\end{ttcolorbox}

\subsection{Baselines}
\label{app:prompts_baselines}

Per conventional established parsing techniques \cite{pymupdf4llm, img2table, llamaindex}, all baseline vision-language and OCR-centric models are evaluated under a uniform prompting setup. This includes DeepSeek-OCR, Gemma-3-27B-IT, GPT-5.2 and Qwen3-VL-8B/32B-Instruct.

\begin{ttcolorbox}[Baseline]
\begin{minipage}{\linewidth}
\begin{lstlisting}[]
Please parse everything in the attached image and output the parsed contents only without anything else.
\end{lstlisting}
\end{minipage}
\end{ttcolorbox}

\subsection{LLM Query Engine}
\label{app:prompts_llm_query_engine}

We use the following prompt to query an LLM along with the attached respective ICL example as context.

\begin{ttcolorbox}[LLM Query Engine]
\begin{minipage}{\linewidth}
\begin{lstlisting}[]
You are a helpful assistant. Use the information from the documents below to answer the question.

{`Retrieved Information'}

Question: {`Question'} 

Answer: 

\end{lstlisting}
\end{minipage}
\end{ttcolorbox}

\newpage
\subsection{Ablation: Structured Representation}
\label{app:prompts_ablation}

To determine the optimal structured representation for tables, we compare our JSON representation (see Appendix \ref{app:prompts_tabrag}) with markdown and natural language text in an ablation study. The prompts used to generate each representation are shown below.

\begin{ttcolorbox}[Ablation: \textsc{TabRAG${_{\text{MD}}}$} (Markdown Table)]
\begin{minipage}{\linewidth}
\begin{lstlisting}[]
You are a precise information parsing and extraction engine. Output ONLY a markdown table of the provided image.

Task: Generate a markdown table with the correct structure, with every single cell accounted for.

Rules:
1. Output ONLY a valid Markdown table. Do not include any other text.
2. Preserve the exact number of rows and columns visible in the image.
3. Preserve unit information in headers if present (e.g., "$ in millions").
4. Transcribe all cell contents as faithfully as possible without inference, normalization, or computation.
5. Repeat values for merged cells so that every row has the same number of columns.
6. If a cell is visually empty, leave it empty.
7. Ignore captions, footnotes, and non-table elements outside the table.

---

{``Self-Generated ICL (Markdown)''}

Now, generate the markdown table ONLY. DO NOT OUTPUT ANYTHING ELSE.
\end{lstlisting}
\end{minipage}
\end{ttcolorbox}

\begin{ttcolorbox}[Ablation: \textsc{TabRAG${_{\text{Text}}}$} (Text Table)]
\begin{minipage}{\linewidth}
\begin{lstlisting}[]
You are an expert document analyst specializing in structured data verbalization. 
Your task is to extract every visible cell from the attached table and convert it into a clear, factual, natural language description.

**Task:** Generate a list of descriptive sentences representing every data point in the table.

**Rules for Generation:**
1.  **Format:** Output a bulleted list of natural language sentences.
2.  **Narrative Flow:** Use conversational connectors like "For...", "the data shows...", "recorded a value of...", or "was found to be...". 
3.  **Hierarchy:** Incorporate nested headers naturally into the sentence (e.g., "Under the 2024 results for Q1, the Revenue was...").
4.  **Preserve Accuracy:** You must mention every specific row, column, and value. Do not skip any cells. Keep all original symbols like ( ), $, and %.
5.  **Contextual Units:** Integrate any units (like "in millions") directly into the description of the value.
6.  **No Rigid Patterns:** Avoid repeating the exact same sentence structure for every line; vary the phrasing slightly to maintain a natural, human-readable tone.

---

{``Self-Generated ICL (Text)''}

Now, extract every cell from the table image into this natural language format. DO NOT OUTPUT ANYTHING ELSE.
\end{lstlisting}
\end{minipage}
\end{ttcolorbox}

\newpage
\begin{ttcolorbox}[Ablation: \textsc{TabRAG${_{\text{MD}}}$} (Markdown Table) Self-Generated ICL]
\begin{minipage}{\linewidth}
\begin{lstlisting}[]
You are a precise information parsing and extraction engine. Output ONLY
A markdown table of the provided image

Task: Generate a markdown table with the correct structure, with every single cell accounted for.

Rules:
1. Output ONLY a valid Markdown table. Do not include any other text.
2. Preserve the exact number of rows and columns visible in the image.
3. Preserve unit information in headers if present (e.g., "$ in millions").
4. Transcribe all cell contents as faithfully as possible without inference, normalization, or computation.
5. Repeat values for merged cells so that every row has the same number of columns.
6. If a cell is visually empty, leave it empty.
7. Ignore captions, footnotes, and non-table elements outside the table.


---

**Example Output 1: Two-level header**

| ($ in millions) | 2024             | 2023             |
|-----------------|------------------|------------------|
|                 | Revenue | Profit | Revenue | Profit |
| Sales           | 1,234   | 400    | 1,200   | 350    |
| Net Income      | (56)    | 80     | -40     | 70     |

---

**Example Output 2: Three-level header**

| ($ in thousands) | 2024                                 | 2023                                |
|------------------|--------------------------------------|-------------------------------------|
|                  | Q1                | Q2               | Q1               | Q2               |
|                  | Revenue | Profit  | Revenue | Profit | Revenue | Profit | Revenue | Profit |
| Product A        | 500     | 120     | 600     | 150    | 450     | 100    | 550     | 140    |
| Product B        | (50)    | 80      | (30)    | 100    | -20     | 60     | 10      | 90     |

---

**Example Output 3: Mixed 1-row, 2-row, and 3-row headers**

| Category              | 2024                                | 2023             | Growth % | Notes    |
|-----------------------|-------------------------------------|------------------|----------|----------|
|                       | Q1               | Q2               | Revenue | Profit |          |          |
|                       | Revenue | Profit | Revenue | Profit |         |        |          |          |
| Sales (Units)         | 1,000   | 300    | 900     | 250    | 1,700   | 550    | 12%      | N/A      |
| Cost ($ in thousands) | (200)   | (50)   | -180    | -40    | (380)   | (90)   | N/A      | Adjusted |

---

Now, generate the markdown table ONLY, do not include any other irrelevant text.
\end{lstlisting}
\end{minipage}
\end{ttcolorbox}

\begin{ttcolorbox}[Ablation: \textsc{TabRAG${_{\text{Text}}}$} (Text Table) Self-Generated ICL]
\begin{minipage}{\linewidth}
\begin{lstlisting}[]
You are an expert document analyst specializing in structured data verbalization. 
Your task is to extract every visible cell from the attached table and convert it into a clear, factual, natural language description.

**Task:** Generate a list of descriptive sentences representing every data point in the table.

**Rules for Generation:**
1.  **Format:** Output a bulleted list of natural language sentences.
2.  **Narrative Flow:** Use conversational connectors like "For...", "the data shows...", "recorded a value of...", or "was found to be...". 
3.  **Hierarchy:** Incorporate nested headers naturally into the sentence (e.g., "Under the 2024 results for Q1, the Revenue was...").
4.  **Preserve Accuracy:** You must mention every specific row, column, and value. Do not skip any cells. Keep all original symbols like ( ), $, and %.
5.  **Contextual Units:** Integrate any units (like "in millions") directly into the description of the value.
6.  **No Rigid Patterns:** Avoid repeating the exact same sentence structure for every line; vary the phrasing slightly to maintain a natural, human-readable tone.

---

**Example 1: Two-level header**

Input:
| ($ in millions) | 2024             | 2023             |
|-----------------|------------------|------------------|
|                 | Revenue | Profit | Revenue | Profit |
| Sales           | 1,234   | 400    | 1,200   | 350    |
| Net Income      | (56)    | 80     | -40     | 70     |

Output:
[
The Sales Revenue in 2024 was 1,234 million dollars.
the Sales Profit in 2024 was 400 million dollars.
the Sales Revenue in 2023 was 1,200 million dollars.
the Sales Profit in 2023 was 350 million dollars.
the Net Income Revenue in 2024 was (56) million dollars.
the Net Income Profit in 2024 was 80 million dollars.
the Net Income Revenue in 2023 was -40 million dollars.
the Net Income Profit in 2023 was 70 million dollars.
]
---

**Example 2: Three-level header**

Input:
| ($ in thousands) | 2024                                 | 2023                                |
|------------------|--------------------------------------|-------------------------------------|
|                  | Q1                | Q2               | Q1               | Q2               |
|                  | Revenue | Profit  | Revenue | Profit | Revenue | Profit | Revenue | Profit |
| Product A        | 500     | 120     | 600     | 150    | 450     | 100    | 550     | 140    |
| Product B        | (50)    | 80      | (30)    | 100    | -20     | 60     | 10      | 90     |

Output:
[
The Q1 Revenue for Product A in 2024 was 500 thousand dollars.
The Q1 Profit for Product A in 2024 was 120 thousand dollars.
The Q2 Revenue for Product A in 2024 was 600 thousand dollars.
The Q2 Profit for Product A in 2024 was 150 thousand dollars.
The Q1 Revenue for Product A in 2023 was 450 thousand dollars.
The Q1 Profit for Product A in 2023 was 100 thousand dollars.
The Q2 Revenue for Product A in 2023 was 550 thousand dollars.
The Q2 Profit for Product A in 2023 was 140 thousand dollars.
The Q1 Revenue for Product B in 2024 was (50) thousand dollars.
The Q1 Profit for Product B in 2024 was 80 thousand dollars.
The Q2 Revenue for Product B in 2024 was (30) thousand dollars.
The Q2 Profit for Product B in 2024 was 100 thousand dollars.
The Q1 Revenue for Product B in 2023 was -20 thousand dollars.
The Q1 Profit for Product B in 2023 was 60 thousand dollars.
The Q2 Revenue for Product B in 2023 was 10 thousand dollars.
The Q2 Profit for Product B in 2023 was 90 thousand dollars.
]

---

**Example 3: Mixed 1-row, 2-row, and 3-row headers**

...

---

Now, extract every cell from the table image into this natural language format. ENSURE ACCURACY IS MAINTAINED.

\end{lstlisting}
\end{minipage}
\end{ttcolorbox}

\newpage
\section{Outputs}
\label{app:outputs}

\subsection{TabRAG: Layout Segmentation}
\label{app:output_layout_detection}
\begin{figure}[H]
    \centering
    \includegraphics[width=0.89\textwidth]{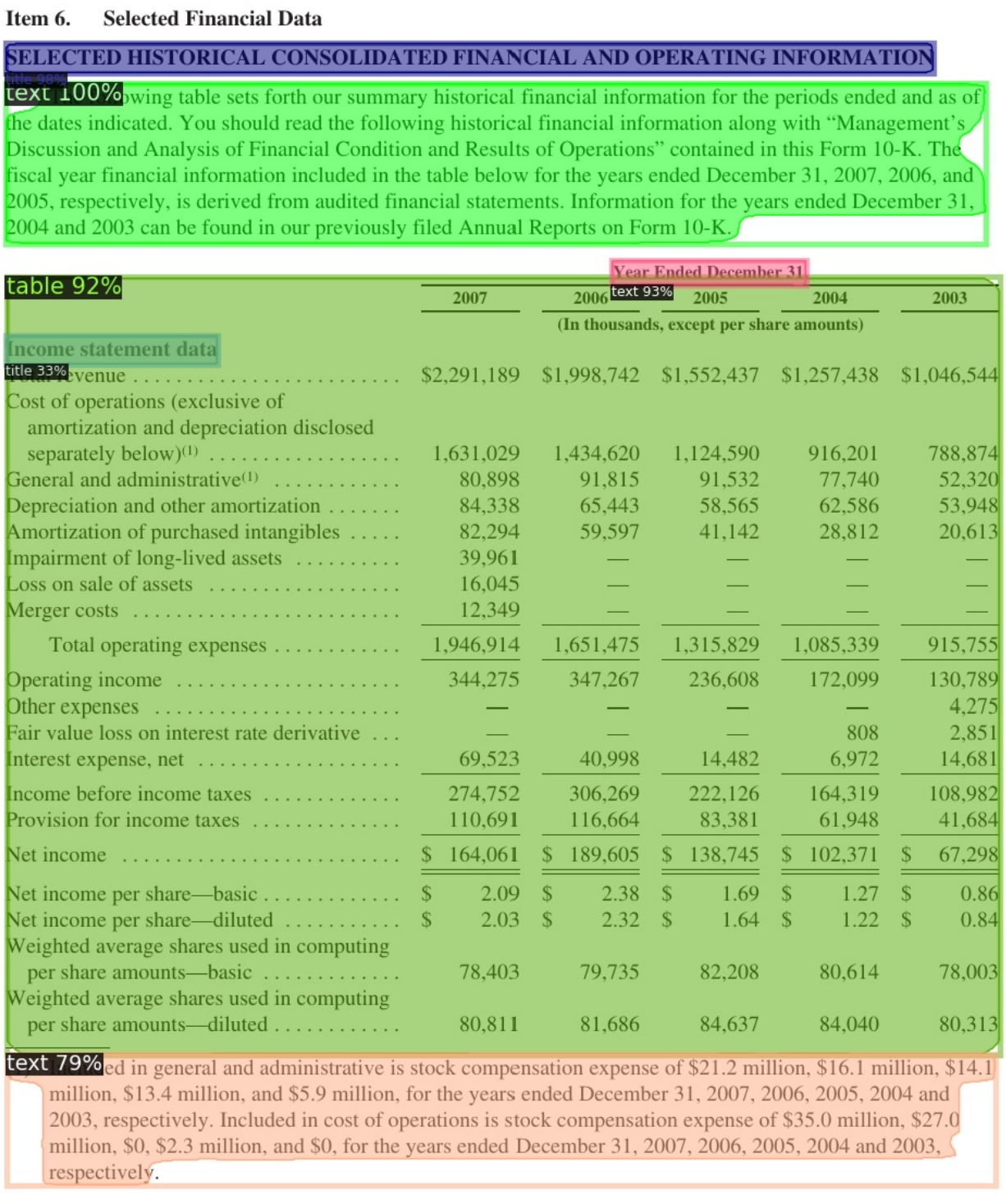}
    \caption{Layout segmentation for a sample document from the ComTQA dataset.}
    \label{fig:layout_example}
    \vspace{-3mm}
\end{figure}

Given a high-resolution document page, our Document Image Transformer-based layout detection model predicts semantically meaningful regions corresponding to various document components such as tables, text blocks, titles, and figures. Each component is associated with a bounding box and a confidence score. As shown in Figure \ref{fig:layout_example}, the model separates the tabular region from the surrounding text blocks and section headers. Each table is detected as a single coherent region containing multiple rows, columns, headers, and numerical entries. This demonstrates the model's ability to preserve fine-grained structural organization that is critical for downstream table reasoning.

\newpage
\subsection{TabRAG}
\label{app:output_tabrag}

In this section, we showcase TabRAG's stage-by-stage outputs using Figure \ref{fig:layout_example}.
\begin{ttcolorbox}[TabRAG: Table]
\begin{minipage}{\linewidth}
\begin{lstlisting}[]
{"row": "Total revenue", "column": "2007", "value": "$2,291,189", "units": "thousand"},
{"row": "Total revenue", "column": "2006", "value": "$1,998,742", "units": "thousand"},
{"row": "Total revenue", "column": "2005", "value": "$1,552,437", "units": "thousand"},
{"row": "Total revenue", "column": "2004", "value": "$1,257,438", "units": "thousand"},
{"row": "Total revenue", "column": "2003", "value": "$1,046,544", "units": "thousand"},
{"row": "Cost of operations (exclusive of amortization and depreciation disclosed separately below)(1)", "column": "2007", 
"value": "1,631,029", "units": "thousand"},
{"row": "Cost of operations (exclusive of amortization and depreciation disclosed separately below)(1)", "column": "2006", 
"value": "1,434,620", "units": "thousand"},
{"row": "Cost of operations (exclusive of amortization and depreciation disclosed separately below)(1)", "column": "2005",
"value": "1,124,590", "units": "thousand"},
{"row": "Cost of operations (exclusive of amortization and depreciation disclosed separately below)(1)", "column": "2004", 
"value": "916,201", "units": "thousand"},
{"row": "Cost of operations (exclusive of amortization and depreciation disclosed separately below)(1)", "column": "2003", 
"value": "788,874", "units": "thousand"},
{"row": "General and administrative(1)", "column": "2007", "value": "80,898", "units": "thousand"},
{"row": "General and administrative(1)", "column": "2006", "value": "91,815", "units": "thousand"},
{"row": "General and administrative(1)", "column": "2005", "value": "91,532", "units": "thousand"},
{"row": "General and administrative(1)", "column": "2004", "value": "77,740", "units": "thousand"},
{"row": "General and administrative(1)", "column": "2003", "value": "52,320", "units": "thousand"},
{"row": "Depreciation and other amortization", "column": "2007", "value": "84,338", "units": "thousand"},
{"row": "Depreciation and other amortization", "column": "2006", "value": "65,443", "units": "thousand"},
{"row": "Depreciation and other amortization", "column": "2005", "value": "58,565", "units": "thousand"},
{"row": "Depreciation and other amortization", "column": "2004", "value": "62,586", "units": "thousand"},
{"row": "Depreciation and other amortization", "column": "2003", "value": "53,948", "units": "thousand"},
{"row": "Amortization of purchased intangibles", "column": "2007", "value": "82,294", "units": "thousand"},
{"row": "Amortization of purchased intangibles", "column": "2006", "value": "59,597", "units": "thousand"},
{"row": "Amortization of purchased intangibles", "column": "2005", "value": "41,142", "units": "thousand"},
{"row": "Amortization of purchased intangibles", "column": "2004", "value": "28,812", "units": "thousand"},
{"row": "Amortization of purchased intangibles", "column": "2003", "value": "20,613", "units": "thousand"},
{"row": "Impairment of long-lived assets", "column": "2007", "value": "39,961", "units": "thousand"},
{"row": "Impairment of long-lived assets", "column": "2006", "value": "-", "units": "thousand"},
{"row": "Impairment of long-lived assets", "column": "2005", "value": "-", "units": "thousand"},
{"row": "Impairment of long-lived assets", "column": "2004", "value": "-", "units": "thousand"},
{"row": "Impairment of long-lived assets", "column": "2003", "value": "-", "units": "thousand"},
{"row": "Loss on sale of assets", "column": "2007", "value": "16,045", "units": "thousand"},
{"row": "Loss on sale of assets", "column": "2006", "value": "-", "units": "thousand"},
{"row": "Loss on sale of assets", "column": "2005", "value": "-", "units": "thousand"},
{"row": "Loss on sale of assets", "column": "2004", "value": "-", "units": "thousand"},
{"row": "Loss on sale of assets", "column": "2003", "value": "-", "units": "thousand"},
{"row": "Merger costs", "column": "2007", "value": "12,349", "units": "thousand"},
{"row": "Merger costs", "column": "2006", "value": "-", "units": "thousand"},
{"row": "Merger costs", "column": "2005", "value": "-", "units": "thousand"},
{"row": "Merger costs", "column": "2004", "value": "-", "units": "thousand"},
{"row": "Merger costs", "column": "2003", "value": "-", "units": "thousand"},
{"row": "Total operating expenses", "column": "2007", "value": "1,946,914", "units": "thousand"},
{"row": "Total operating expenses", "column": "2006", "value": "1,651,475", "units": "thousand"},
{"row": "Total operating expenses", "column": "2005", "value": "1,315,829", "units": "thousand"},
{"row": "Total operating expenses", "column": "2004", "value": "1,085,339", "units": "thousand"},
{"row": "Total operating expenses", "column": "2003", "value": "915,755", "units": "thousand"},
{"row": "Operating income", "column": "2007", "value": "344,275", "units": "thousand"},
{"row": "Operating income", "column": "2006", "value": "347,267", "units": "thousand"},
{"row": "Operating income", "column": "2005", "value": "236,608", "units": "thousand"},
{"row": "Operating income", "column": "2004", "value": "172,099", "units": "thousand"},
{"row": "Operating income", "column": "2003", "value": "130,789", "units": "thousand"},
{"row": "Other expenses", "column": "2007", "value": "-", "units": "thousand"},
{"row": "Other expenses", "column": "2006", "value": "-", "units": "thousand"},
{"row": "Other expenses", "column": "2005", "value": "-", "units": "thousand"},
{"row": "Other expenses", "column": "2004", "value": "-", "units": "thousand"},
{"row": "Other expenses", "column": "2003", "value": "4,275", "units": "thousand"},
{"row": "Fair value loss on interest rate derivative", "column": "2007", "value": "-", "units": "thousand"},
{"row": "Fair value loss on interest rate derivative", "column": "2006", "value": "-", "units": "thousand"},
{"row": "Fair value loss on interest rate derivative", "column": "2005", "value": "-", "units": "thousand"},
{"row": "Fair value loss on interest rate derivative", "column": "2004", "value": "808", "units": "thousand"},
{"row": "Fair value loss on interest rate derivative", "column": "2003", "value": "2,851", "units": "thousand"},
{"row": "Interest expense, net", "column": "2007", "value": "69,523", "units": "thousand"},
{"row": "Interest expense, net", "column": "2006", "value": "40,998", "units": "thousand"},
{"row": "Interest expense, net", "column": "2005", "value": "14,482", "units": "thousand"},
{"row": "Interest expense, net", "column": "2004", "value": "6,972", "units": "thousand"},
{"row": "Interest expense, net", "column": "2003", "value": "14,681", "units": "thousand"},
{"row": "Income before income taxes", "column": "2007", "value": "274,752", "units": "thousand"},
{"row": "Income before income taxes", "column": "2006", "value": "306,269", "units": "thousand"},
{"row": "Income before income taxes", "column": "2005", "value": "222,126", "units": "thousand"},
{"row": "Income before income taxes", "column": "2004", "value": "164,319", "units": "thousand"},
{"row": "Income before income taxes", "column": "2003", "value": "108,982", "units": "thousand"},
{"row": "Provision for income taxes", "column": "2007", "value": "110,691", "units": "thousand"},
{"row": "Provision for income taxes", "column": "2006", "value": "116,664", "units": "thousand"},
{"row": "Provision for income taxes", "column": "2005", "value": "83,381", "units": "thousand"},
{"row": "Provision for income taxes", "column": "2004", "value": "61,948", "units": "thousand"},
{"row": "Provision for income taxes", "column": "2003", "value": "41,684", "units": "thousand"},
{"row": "Net income", "column": "2007", "value": "$ 164,061", "units": "thousand"},
{"row": "Net income", "column": "2006", "value": "$ 189,605", "units": "thousand"},
\end{lstlisting}
\end{minipage}
\end{ttcolorbox}

\begin{ttcolorbox}[]
\begin{minipage}{\linewidth}
\begin{lstlisting}[]
{"row": "Net income", "column": "2005", "value": "$ 138,745", "units": "thousand"},
{"row": "Net income", "column": "2004", "value": "$ 102,371", "units": "thousand"},
{"row": "Net income", "column": "2003", "value": "$ 67,298", "units": "thousand"},
{"row": "Net income per share-basic", "column": "2007", "value": "$ 2.09", "units": "null"},
{"row": "Net income per share-basic", "column": "2006", "value": "$ 2.38", "units": "null"},
{"row": "Net income per share-basic", "column": "2005", "value": "$ 1.69", "units": "null"},
{"row": "Net income per share-basic", "column": "2004", "value": "$ 1.27", "units": "null"},
{"row": "Net income per share-basic", "column": "2003", "value": "$ 0.86", "units": "null"},
{"row": "Net income per share-diluted", "column": "2007", "value": "$ 2.03", "units": "null"},
{"row": "Net income per share-diluted", "column": "2006", "value": "$ 2.32", "units": "null"},
{"row": "Net income per share-diluted", "column": "2005", "value": "$ 1.64", "units": "null"},
{"row": "Net income per share-diluted", "column": "2004", "value": "$ 1.22", "units": "null"},
{"row": "Net income per share-diluted", "column": "2003", "value": "$ 0.84", "units": "null"},
{"row": "Weighted average shares used in computing per share amounts-basic", "column": "2007", "value": "78,403", "units": "null"},
{"row": "Weighted average shares used in computing per share amounts-basic", "column": "2006", "value": "79,735", "units": "null"},
{"row": "Weighted average shares used in computing per share amounts-basic", "column": "2005", "value": "82,208", "units": "null"},
{"row": "Weighted average shares used in computing per share amounts-basic", "column": "2004", "value": "80,614", "units": "null"},
{"row": "Weighted average shares used in computing per share amounts-basic", "column": "2003", "value": "78,003", "units": "null"},
{"row": "Weighted average shares used in computing per share amounts-diluted", "column": "2007", "value": "80,811", 
"units": "null"},
{"row": "Weighted average shares used in computing per share amounts-diluted", "column": "2006", "value": "81,686", 
"units": "null"},
{"row": "Weighted average shares used in computing per share amounts-diluted", "column": "2005", "value": "84,637", 
"units": "null"},
{"row": "Weighted average shares used in computing per share amounts-diluted", "column": "2004", "value": "84,040", 
"units": "null"},
{"row": "Weighted average shares used in computing per share amounts-diluted", "column": "2003", "value": "80,313", 
"units": "null"}
\end{lstlisting}
\end{minipage}
\end{ttcolorbox}

\newpage
\begin{ttcolorbox}[TabRAG: Title]
\begin{minipage}{\linewidth}
\begin{lstlisting}[]
SELECTED HISTORICAL CONSOLIDATED FINANCIAL AND OPERATING INFORMATION
Income statement data
\end{lstlisting}
\end{minipage}
\end{ttcolorbox}

\begin{ttcolorbox}[TabRAG: Text]
\begin{minipage}{\linewidth}
\begin{lstlisting}[]
The following table sets forth our summary historical financial information for the periods ended and as of the dates indicated. 
You should read the following historical financial information along with "Management's Discussion and Analysis of Financial 
Condition and Results of Operations" contained in this Form 10-K. The fiscal year financial information included in the table below 
for the years ended December 31, 2007, 2006, and 2005, respectively, is derived from audited financial statements. Information for 
the years ended December 31, 2004 and 2003 can be found in our previously filed Annual Reports on Form 10-K.
Year Ended December 31,
\end{lstlisting}
\end{minipage}
\end{ttcolorbox}

\begin{ttcolorbox}[TabRAG: Page Overview]
\begin{minipage}{\linewidth}
\begin{lstlisting}[]
Item 6. Selected Financial Data
SELECTED HISTORICAL CONSOLIDATED FINANCIAL AND OPERATING INFORMATION
The following table sets forth our summary historical financial information for the periods ended and as of the dates indicated. 
You should read the following historical financial information along with "Management's Discussion and Analysis of Financial 
Condition and Results of Operations" contained in this Form 10-K. The fiscal year financial information included in the table below 
for the years ended December 31, 2007, 2006, and 2005, respectively, is derived from audited financial statements. Information for 
the years ended December 31, 2004 and 2003 can be found in our previously filed Annual Reports on Form 10-K.
Year Ended December 31,
2007     2006     2005     2004     2003
(In thousands, except per share amounts)
Income statement data
Total revenue  $2,291,189  $1,998,742  $1,552,437  $1,257,438  $1,046,544
Cost of operations (exclusive of
amortization and depreciation disclosed
separately below)(1)  1,631,029  1,434,620  1,124,590  916,201  788,874
General and administrative(1)  80,898  91,815  91,532  77,740  52,320
Depreciation and other amortization  84,338  65,443  58,565  62,586  53,948
Amortization of purchased intangibles  82,294  59,597  41,142  28,812  20,613
Impairment of long-lived assets  39,961  -  -  -  -
Loss on sale of assets  16,045  -  -  -  -
Merger costs  12,349  -  -  -  -
Total operating expenses  1,946,914  1,651,475  1,315,829  1,085,339  915,755
Operating income  344,275  347,267  236,608  172,099  130,789
Other expenses  -  -  -  -  4,275
Fair value loss on interest rate derivative  -  -  -  808  2,851
Interest expense, net  69,523  40,998  14,482  6,972  14,681
Income before income taxes  274,752  306,269  222,126  164,319  108,982
Provision for income taxes  110,691  116,664  83,381  61,948  41,684
Net income  $ 164,061  $ 189,605  $ 138,745  $ 102,371  $ 67,298
Net income per share-basic  $ 2.09  $ 2.38  $ 1.69  $ 1.27  $ 0.86
Net income per share-diluted  $ 2.03  $ 2.32  $ 1.64  $ 1.22  $ 0.84
Weighted average shares used in computing
per share amounts-basic  78,403  79,735  82,208  80,614  78,003
Weighted average shares used in computing
per share amounts-diluted  80,811  81,686  84,637  84,040  80,313
(1) Included in general and administrative is stock compensation expense of $21.2 million, $16.1 million, $14.1 million, $13.4 
million, and $5.9 million, for the years ended December 31, 2007, 2006, 2005, 2004 and 2003, respectively. Included in cost of 
operations is stock compensation expense of $35.0 million, $27.0 million, $0, $2.3 million, and $0, for the years ended 
December 31, 2007, 2006, 2005, 2004 and 2003, respectively.
\end{lstlisting}
\end{minipage}
\end{ttcolorbox}

\newpage
\subsection{Ablation: Structured Representation (Case Study)}
\label{app:output_ablation-struc-rep}

\begin{figure}[H]
    \centering
    \includegraphics[width=0.9\textwidth]{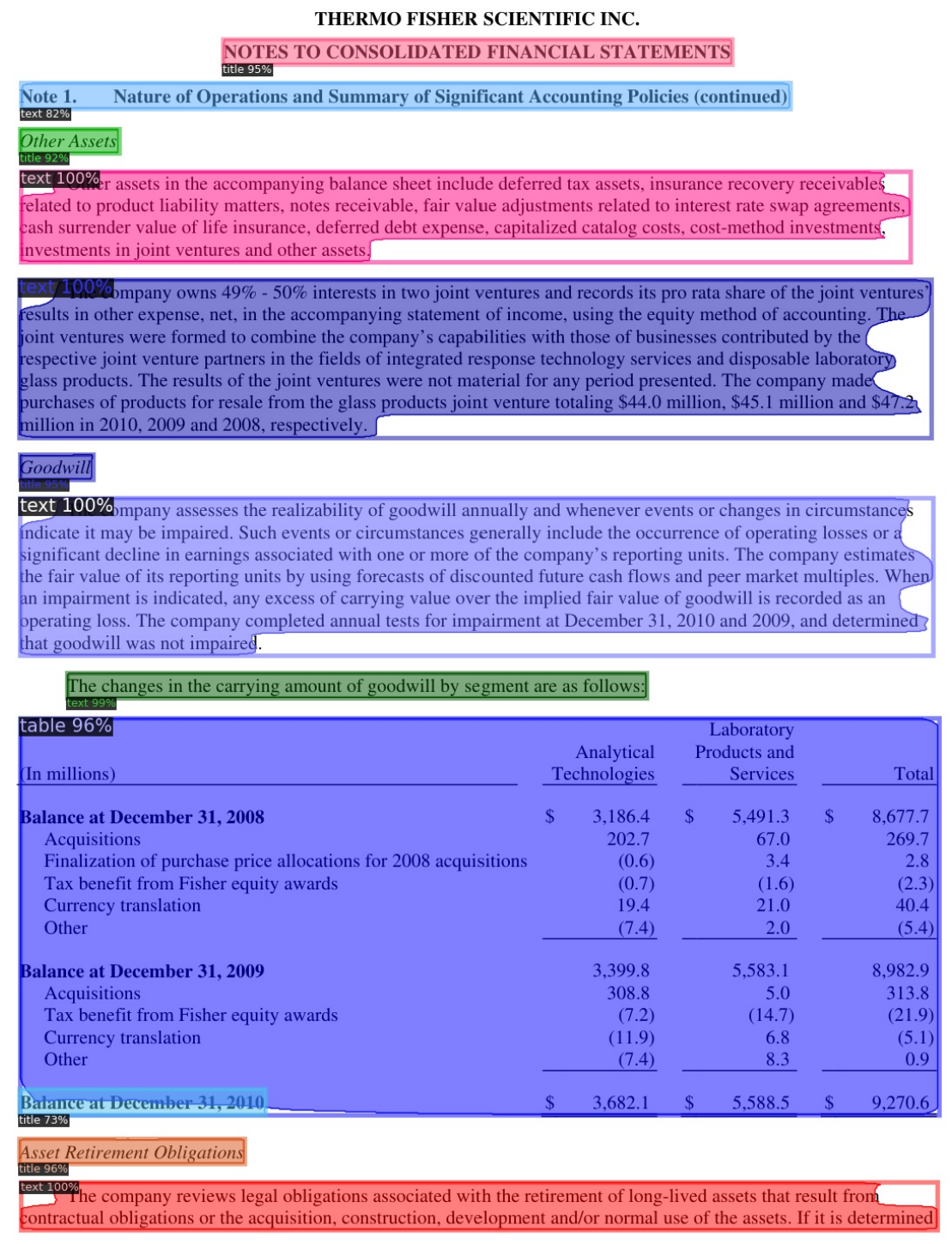}
    \caption{An example layout segmentation output from ComTQA.}
    \label{fig:layout_example_comtqa}
    \vspace{-3mm}
\end{figure}

In this section, we perform a case study by providing a visual comparison of the outputs generated by each structured representation method used in our ablation experiments. We use a sample document from the ComTQA dataset in Figure \ref{fig:layout_example_comtqa}.

The following illustrates the constructed representations.

\begin{ttcolorbox}[\textsc{TabRAG}$_\text{JSON}$ (Table)]
\begin{minipage}{\linewidth}
\begin{lstlisting}[]
{"row": "Balance at December 31, 2008", "column": "Analytical Technologies", "value": "$ 3,186.4", "units": "million"}
{"row": "Balance at December 31, 2008", "column": "Laboratory Products and Services", "value": "$ 5,491.3", "units": "million"}
{"row": "Balance at December 31, 2008", "column": "Total", "value": "$ 8,677.7", "units": "million"}
{"row": "Acquisitions", "column": "Analytical Technologies", "value": "202.7", "units": "million"}
{"row": "Acquisitions", "column": "Laboratory Products and Services", "value": "67.0", "units": "million"}
{"row": "Acquisitions", "column": "Total", "value": "269.7", "units": "million"}
{"row": "Finalization of purchase price allocations for 2008 acquisitions", "column": "Analytical Technologies", "value": "(0.6)", 
"units": "million"}
{"row": "Finalization of purchase price allocations for 2008 acquisitions", "column": "Laboratory Products and Services", "value": 
"3.4", "units": "million"}
{"row": "Finalization of purchase price allocations for 2008 acquisitions", "column": "Total", "value": "2.8", "units": "million"}
{"row": "Tax benefit from Fisher equity awards", "column": "Analytical Technologies", "value": "(0.7)", "units": "million"}
{"row": "Tax benefit from Fisher equity awards", "column": "Laboratory Products and Services", "value": "(1.6)", 
"units": "million"}
{"row": "Tax benefit from Fisher equity awards", "column": "Total", "value": "(2.3)", "units": "million"}
{"row": "Currency translation", "column": "Analytical Technologies", "value": "19.4", "units": "million"}
{"row": "Currency translation", "column": "Laboratory Products and Services", "value": "21.0", "units": "million"}
{"row": "Currency translation", "column": "Total", "value": "40.4", "units": "million"}
{"row": "Other", "column": "Analytical Technologies", "value": "(7.4)", "units": "million"}
{"row": "Other", "column": "Laboratory Products and Services", "value": "2.0", "units": "million"}
{"row": "Other", "column": "Total", "value": "(5.4)", "units": "million"}
{"row": "Balance at December 31, 2009", "column": "Analytical Technologies", "value": "3,399.8", "units": "million"}
{"row": "Balance at December 31, 2009", "column": "Laboratory Products and Services", "value": "5,583.1", "units": "million"}
{"row": "Balance at December 31, 2009", "column": "Total", "value": "8,982.9", "units": "million"}
{"row": "Acquisitions", "column": "Analytical Technologies", "value": "308.8", "units": "million"}
{"row": "Acquisitions", "column": "Laboratory Products and Services", "value": "5.0", "units": "million"}
{"row": "Acquisitions", "column": "Total", "value": "313.8", "units": "million"}
{"row": "Tax benefit from Fisher equity awards", "column": "Analytical Technologies", "value": "(7.2)", "units": "million"}
{"row": "Tax benefit from Fisher equity awards", "column": "Laboratory Products and Services", "value": "(14.7)", 
"units": "million"}
{"row": "Tax benefit from Fisher equity awards", "column": "Total", "value": "(21.9)", "units": "million"}
{"row": "Currency translation", "column": "Analytical Technologies", "value": "(11.9)", "units": "million"}
{"row": "Currency translation", "column": "Laboratory Products and Services", "value": "6.8", "units": "million"}
{"row": "Currency translation", "column": "Total", "value": "(5.1)", "units": "million"}
{"row": "Other", "column": "Analytical Technologies", "value": "(7.4)", "units": "million"}
{"row": "Other", "column": "Laboratory Products and Services", "value": "8.3", "units": "million"}
{"row": "Other", "column": "Total", "value": "0.9", "units": "million"}
{"row": "Balance at December 31, 2010", "column": "Analytical Technologies", "value": "$ 3,682.1", "units": "million"}
{"row": "Balance at December 31, 2010", "column": "Laboratory Products and Services", "value": "$ 5,588.5", "units": "million"}
{"row": "Balance at December 31, 2010", "column": "Total", "value": "$ 9,270.6", "units": "million"}
\end{lstlisting}
\end{minipage}
\end{ttcolorbox}

\newpage
\begin{ttcolorbox}[\textsc{TabRAG${_{\text{MD}}}$} (Table)]
\begin{minipage}{\linewidth}
\begin{lstlisting}[]
| | Analytical Technologies | Laboratory Products and Services | Total |\n|---|---|---|---|\n| (In millions) | | | |\n| Balance at 
December 31, 2008 | $ 3,186.4 | $ 5,491.3 | $ 8,677.7 |\n| Acquisitions | 202.7 | 67.0 | 269.7 |\n| Finalization of purchase 
price allocations for 2008 acquisitions | (0.6) | 3.4 | 2.8 |\n| Tax benefit from Fisher equity awards | (0.7) | (1.6) | 
(2.3) |\n| Currency translation | 19.4 | 21.0 | 40.4 |\n| Other | (7.4) | 2.0 | (5.4) |\n| Balance at December 31, 2009 | 
3,399.8 | 5,583.1 | 8,982.9 |\n| Acquisitions | 308.8 | 5.0 | 313.8 |\n| Tax benefit from Fisher equity awards | (7.2) | 
(14.7) | (21.9) |\n| Currency translation | (11.9) | 6.8 | (5.1) |\n| Other | (7.4) | 8.3 | 0.9 |\n| Balance at December 31, 
2010 | $ 3,682.1 | $ 5,588.5 | $ 9,270.6 |\nBalance at December 31, 2010}
\end{lstlisting}
\end{minipage}
\end{ttcolorbox}

\begin{ttcolorbox}[\textsc{TabRAG${_{\text{Text}}}$} (Table)]
\begin{minipage}{\linewidth}
\begin{lstlisting}[]
The Balance at December 31, 2008 for Analytical Technologies was $3,186.4 million.
The Balance at December 31, 2008 for Laboratory Products and Services was $5,491.3 million.
The Balance at December 31, 2008 for the Total was $8,677.7 million.
The Acquisitions for Analytical Technologies in 2008 were $202.7 million.
The Acquisitions for Laboratory Products and Services in 2008 were $67.0 million.
The Acquisitions for the Total in 2008 were $269.7 million.
The Finalization of purchase price allocations for 2008 acquisitions for Analytical Technologies was $(0.6) million.
The Finalization of purchase price allocations for 2008 acquisitions for Laboratory Products and Services was $3.4 million.
The Finalization of purchase price allocations for 2008 acquisitions for the Total was $2.8 million.
The Tax benefit from Fisher equity awards for Analytical Technologies in 2008 was $(0.7) million.
The Tax benefit from Fisher equity awards for Laboratory Products and Services in 2008 was $(1.6) million.
The Tax benefit from Fisher equity awards for the Total in 2008 was $(2.3) million.
The Currency translation for Analytical Technologies in 2008 was $19.4 million.
The Currency translation for Laboratory Products and Services in 2008 was $21.0 million.
The Currency translation for the Total in 2008 was $40.4 million.
The Other for Analytical Technologies in 2008 was $(7.4) million.
The Other for Laboratory Products and Services in 2008 was $2.0 million.
The Other for the Total in 2008 was $(5.4) million.
The Balance at December 31, 2009 for Analytical Technologies was $3,399.8 million.
The Balance at December 31, 2009 for Laboratory Products and Services was $5,583.1 million.
The Balance at December 31, 2009 for the Total was $8,982.9 million.
The Acquisitions for Analytical Technologies in 2009 were $308.8 million.
The Acquisitions for Laboratory Products and Services in 2009 were $5.0 million.
The Acquisitions for the Total in 2009 were $313.8 million.
The Tax benefit from Fisher equity awards for Analytical Technologies in 2009 was $(7.2) million.
The Tax benefit from Fisher equity awards for Laboratory Products and Services in 2009 was $(14.7) million.
The Tax benefit from Fisher equity awards for the Total in 2009 was $(21.9) million.
The Currency translation for Analytical Technologies in 2009 was $(11.9) million.
The Currency translation for Laboratory Products and Services in 2009 was $6.8 million.
The Currency translation for the Total in 2009 was $(5.1) million.
The Other for Analytical Technologies in 2009 was $(7.4) million.
The Other for Laboratory Products and Services in 2009 was $8.3 million.
The Other for the Total in 2009 was $0.9 million.
The Balance at December 31, 2010 for Analytical Technologies was $3,682.1 million.
The Balance at December 31, 2010 for Laboratory Products and Services was $5,588.5 million.
The Balance at December 31, 2010 for the Total was $9,270.6 million.
\end{lstlisting}
\end{minipage}
\end{ttcolorbox}

\newpage
We evaluate the generation performance using the representations with ComTQA's QA set. For this document, it contains the following QAs:

\begin{itemize}
    \item Q1: What is the difference between the total amount of assets at the end of 2008 and the total amount of assets at the end of 2009?
    \item A1: \$305.2 million
    \item Q2: What is the difference between the total amount of assets at the end of 2009 and the total amount of assets at the end of 2010?
    \item A2: \$287.7 million
\end{itemize}

The following outline the output from the various representations.
\begin{ttcolorbox}[\textsc{TabRAG}$_\text{JSON}$ (Generation Output)]
\begin{minipage}{\linewidth}
\begin{lstlisting}[]
Question 1 (Correct):
The difference between the total amount of assets at the end of 2008 and the end of 2009 is calculated by subtracting the total 
assets at the end of 2008 from the total assets at the end of 2009. 
Total assets at the end of 2008: $8,677.7 million  
Total assets at the end of 2009: $8,982.9 million  
Difference: $8,982.9 million - $8,677.7 million = $305.2 million

Question 2 (Correct): The difference between the total amount of assets at the end of 2009 and the end of 2010 is calculated by 
subtracting the total assets at the end of 2009 from the total assets at the end of 2010. 

Total assets at the end of 2009: $8,982.9 million  
Total assets at the end of 2010: $9,270.6 million  
Difference: $9,270.6 million - $8,982.9 million = $287.7 million
\end{lstlisting}
\end{minipage}
\end{ttcolorbox}

\begin{ttcolorbox}[\textsc{TabRAG}$_\text{MD}$ (Generation Output)]
\begin{minipage}{\linewidth}
\begin{lstlisting}[]
Question 1 (Incorrect):
The provided information does not include the total amount of assets at the end of 2008 and 2009, making it impossible to calculate 
the difference. Therefore, the answer is not available from the given data.

Question 2 (Incorrect): 
The provided information does not include the total amount of assets at the end of 2009 and 2010, making it impossible to calculate 
the difference between the two amounts. The data available only provides details about goodwill and other specific assets, not the 
overall total assets for the years 2009 and 2010.
\end{lstlisting}
\end{minipage}
\end{ttcolorbox}

\begin{ttcolorbox}[\textsc{TabRAG}$_\text{Text}$ (Generation Output)]
\begin{minipage}{\linewidth}
\begin{lstlisting}[]
Question 1 (Incorrect): 
The difference between the total amount of assets at the end of 2008 and the end of 2009 cannot be directly calculated from the 
provided information. The text provided focuses on specific components of assets, such as goodwill, joint ventures, and other 
assets, but does not provide the total assets for the end of 2008 or 2009. To calculate the difference, the total assets for both 
years would need to be known, which are not explicitly provided in the given text.

Question 2 (Incorrect): 
The difference between the total amount of assets at the end of 2009 and the end of 2010 cannot be directly calculated from the 
provided information. The text provided focuses on specific components of assets, such as goodwill, joint ventures, and other 
assets, but does not provide the total assets for the end of 2009 or 2010. To calculate the difference, the total assets for both 
years would need to be known, which are not explicitly provided in the text.

\end{lstlisting}
\end{minipage}
\end{ttcolorbox}

\end{document}